\documentclass[10pt,twocolumn,letterpaper]{article}

\usepackage{iccv}      

\definecolor{iccvblue}{rgb}{0.21,0.49,0.74}
\usepackage[pagebackref,breaklinks,colorlinks,allcolors=iccvblue]{hyperref}
\usepackage{multirow}

\def\paperID{2925} 
\def\confName{ICCV}
\def\confYear{2025}
\usepackage{amsmath}
\usepackage{algorithm}
\usepackage{algorithmicx}
\usepackage{algpseudocode}
\usepackage{diagbox}

\title{SyncDiff: Synchronized Motion Diffusion for Multi-Body\\Human-Object Interaction Synthesis}

\author{
Wenkun He\textsuperscript{1,2},~
Yun Liu\textsuperscript{1,2,3},~
Ruitao Liu\textsuperscript{1},~
Li Yi\textsuperscript{\textdagger,1,2,3}
\smallskip\\
\textsuperscript{1}Tsinghua University~~~
\textsuperscript{2}Shanghai Qi Zhi Institute~~~
\textsuperscript{3}Shanghai Artificial Intelligence Laboratory\\
\url{https://syncdiff.github.io/}
}

\begin{document}

\twocolumn[{
\renewcommand\twocolumn[1][]{#1}
\maketitle
\begin{center}
    \captionsetup{type=figure}
    \includegraphics[width=1.0\textwidth]{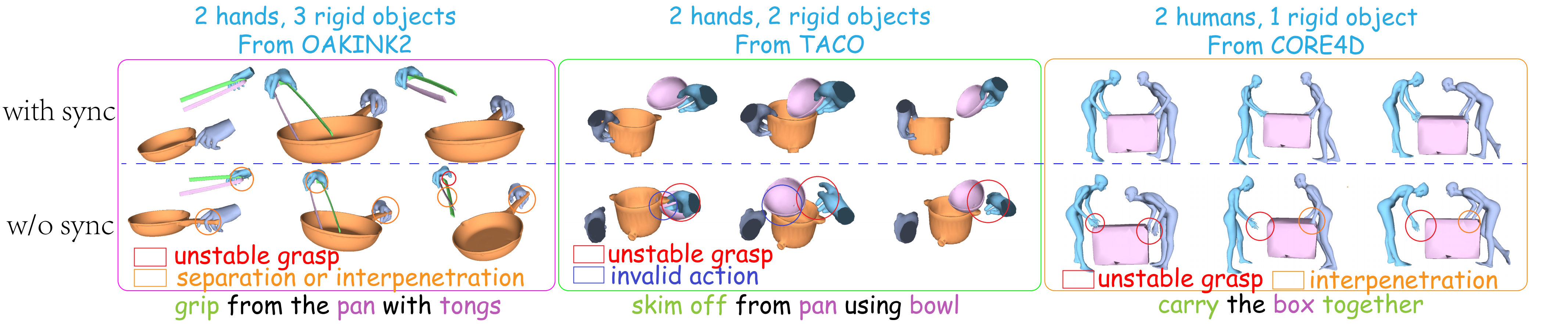}
    \caption{
    SyncDiff is a unified framework synthesizing synchronized multi-body interaction motions with any number of hands, humans, and rigid objects. In SyncDiff, we introduce two novel multi-body motion synchronization mechanisms, namely the alignment scores for training and explicit synchronization strategy in inference. With these mechanisms, the synthesized results can effectively prevent interpenetration, contact loss, or asynchronous human-object interactions in various scenarios, as shown in the above figure.
    }
    \label{fig:teaser}
\end{center}
}]

\begin{abstract}
Synthesizing realistic human-object interaction motions is a critical problem in VR/AR and human animation. Unlike the commonly studied scenarios involving a single human or hand interacting with one object, we address a more generic multi-body setting with arbitrary numbers of humans, hands, and objects. The high correlations and mutual influences among bodies leads to two major challenges, for which we propose solutions. First, to satisfy the high demands for synchronization of different body motions, we mathematically derive a new set of alignment scores during the training process, and use maximum likelihood sampling on a dynamic graphical model for explicit synchronization during inference. Second, the high-frequency interactions between objects are often overshadowed by the large-scale low-frequency movements. To address this, we introduce frequency decomposition and explicitly represent high-frequency components in the frequency domain. Extensive experiments across five datasets with various multi-body configurations demonstrate the superiority of SyncDiff over existing state-of-the-art motion synthesis methods.
\end{abstract}
\section{Introduction}
\label{sec:1}

In daily life, humans frequently manipulate objects to complete tasks, often using both hands or collaborating with others. For instance, an individual might use both hands to set up a fallen chair, hold a brush in one hand to scrub a bowl held in the other, or work with another person to lift and position a heavy object. These are examples of multi-body human-object interactions (HOI)~\cite{li2023objectmotionguidedhuman,christen2024diffh2odiffusionbasedsynthesishandobject,li2023controllable}, where ``body'' can refer to objects, hands, or humans. Synthesizing such interactions has prominent applications in VR/AR, human animation, and robot learning~\cite{wang2024genh2rlearninggeneralizablehumantorobot}.

The primary challenge in multi-body HOI synthesis is capturing the complex joint distribution of body motions and ensuring that the individual body motions are not only synchronized but also mutually aligned with meaningful interaction semantics. This challenge intensifies as the number of bodies increases with high-order motion relationships. Existing works only focus on some specific configurations with limited body numbers, such as a single hand interacting with an object~\cite{wang2024genh2rlearninggeneralizablehumantorobot}, a person engaging with an object~\cite{li2023objectmotionguidedhuman, MDM}, or two hands manipulating a single item~\cite{shimada2023macsmassconditioned3d, christen2024diffh2odiffusionbasedsynthesishandobject}, where motion trajectories are often simple and without too many semantics. These works introduce configuration-specific priors and are restricted to particular setups, leading to a strong demand for a generic method that could handle the series of multi-body HOI synthesis problems in a unified manner without restrictions for body numbers, and with more complex motion distributions.

A straightforward method to promote synchronization is treating the motions of all individual bodies as a high-dimensional representation and using a single diffusion model to depict its distribution~\cite{christen2024diffh2odiffusionbasedsynthesishandobject}. However, this diffusion model only estimates data sample scores, and only implicitly depicts the correlations across individual motions. To further promote the alignment between individual body motions with limited data amount, analogous to the data sample scores guiding data denoising and reconstruction, we need a set of alignment scores to promote motion synchronization. Second, with estimated sample and alignment scores, in inference, we can jointly optimize sample and alignment likelihoods by formulating multi-body HOI synthesis as a motion synchronization problem within a graphical model, where nodes represent individual body motions and edges capture the relative motions between body pairs.

In practice, we often observe high-frequency, small-amplitude vibrations with semantics be overshadowed by low-frequency, large-scale movements (e.g., A brush only contacts the teapot without effective frictions between them). To solve this, we propose frequency decomposition, which factorizes all individual and relative motions into low and high frequency components, and supervises them in the time domain and frequency domain separately.

To realize the ideas above, we design Synchronized Motion Diffusion (SyncDiff) for generic multi-body HOI synthesis. SyncDiff is a diffusion model defined on the graphical model above, operating on a high-order motion representation consisting of all individual motions and relative motions, which is the first unified framework for multi-body HOI synthesis of any body count. Our detailed contributions include: 1) For better synchronization, we derive a set of \textbf{alignment scores} and the corresponding loss term for training. 2) At inference time, by leveraging both sample and alignment scores, we discover a simple \textbf{explicit synchronization} strategy, which is equivalent to maximum likelihood inference with a mathematical justification. 3) We \textbf{decompose} all individual and relative motions based on \textbf{frequency}, to better model high-frequency components with semantics. 4) Extensive experiments on \textbf{five datasets} demonstrate the superior human-object contact quality of our method, and the semantic recognition accuracy rate surpasses the sota methods by an average of \textbf{over 15\%}.
\section{Related Works}
\label{sec:related}

\subsection{Human-object Interaction Motion Synthesis}
\label{sec:2.1}


Synthesizing interactive motions between humans and objects has been an emerging research field supported by extensive HOI datasets ~\cite{brahmbhatt2019contactdbanalyzingpredictinggrasp, brahmbhatt2020contactpose, chao2021dexycbbenchmarkcapturinghand, fan2023arcticdatasetdexterousbimanual, hampali2020honnotatemethod3dannotation, kwon2021h2ohandsmanipulatingobjects, liu2022hoi4d, taheri2020grab, yang2022oakinklargescaleknowledgerepository, krebs2021kit, zhan2024oakink2datasetbimanualhandsobject, liu2024tacobenchmarkinggeneralizablebimanual, zhang2024core4d4dhumanobjecthumaninteraction, xu2024interx,bhatnagar2022behave,zhang2025motionxlargescalemultimodal3d}. A branch of works~\cite{hassan2021stochasticsceneawaremotionprediction, mir2023generatingcontinualhumanmotion, wang2021synthesizinglongterm3dhuman, wang2021sceneawaregenerativenetworkhuman, zhao2023synthesizingdiversehumanmotions} aims at synthesizing human bodies interacting with objects, while others~\cite{rajeswaran2018learningcomplexdexterousmanipulation, garciahernando2020physicsbaseddexterousmanipulationsestimated, mandikal2021learningdexterousgraspingobjectcentric, qin2022dexmvimitationlearningdexterous, wang2024genh2rlearninggeneralizablehumantorobot, zhou2024learningdiversebimanualdexterous,zhang2024manidext,cha2024text2hoi,ishant2025controllablehandgraspgeneration} focus on hand-object interactions for learning more dexterous manipulation behaviors.
For synthesizing human-object interaction motions, early works~\cite{COINS,IMoS,zhang2022couch,taheri2022goal,qi2025humangraspgenerationrigid} propose to leverage conditional variational auto-encoders~\cite{kingma2013auto,sohn2015learning} to model distributions of human-object interaction motions and enable generalization ability, meanwhile using CLIP~\cite{radford2021learningtransferablevisualmodels} to encode language features for text-based control.
With the tremendous progress of diffusion models~\cite{ho2020denoisingdiffusionprobabilisticmodels}, diffusion-based methods~\cite{MDM,li2023objectmotionguidedhuman,li2023controllable,xu2023interdiff,peng2024hoidifftextdrivensynthesis3d,wu2024human,CG-HOI,wu2024thor, cong2025semgeomodynamiccontextualhuman,wang2024regiongraspnoveltaskcontact} have been widely proposed for superior generation qualities. Focusing on multi-person and object collaboration synthesis, recent approaches~\cite{shafir2023humanmotiondiffusiongenerative,Liang_2024,peng2024hoidifftextdrivensynthesis3d,daiya2024collagecollaborativehumanagentinteraction,tanaka2023role,xu2024ReGenNet,zhang2024oodhoitextdriven3dwholebody} enhance feature exchange among different entities in diffusion steps for better cross-entity synchronization.
For synthesizing hand-object interaction motions, one solution is to train dexterous hand agents~\cite{D-Grasp, Chen_2023,zhang2024artigraspphysicallyplausiblesynthesis,zhang2025graspxl,wan2023unidexgrasp++,xu2025intermimicuniversalwholebodycontrol} using physical simulations~\cite{todorov2012mujoco,isaacgym} and reinforcement learning~\cite{SAC,PPO}.
Another solution that is similar to ours is fully data-driven. Early methodologies~\cite{DexRepNet,ContactGen,CPF,manipnet,zheng2023cams,TOCH} apply representation techniques such as contact map or reference grasp, which better model the contact between hand joints or hand surface points and the relevant local regions of the object. Several recent works~\cite{christen2023learning,shimada2023macsmassconditioned3d,zhang2024manidext,cha2024text2hoi,zhang2024bimart} have shifted the focus to multi-hand collaboration, while few works have addressed interactions involving more than one object.
Compared to the above methods, our framework can handle both human-object and hand-object interactions, allowing any number of bodies in the scene, without the need for any delicately designed structures like contact guidance.




\subsection{Injecting External Knowledge into Diffusion Models}
\label{sec:2.2}
Benefitting from high generation qualities, diffusion models~\cite{ho2020denoisingdiffusionprobabilisticmodels,ddim} are widely adopted in versatile synthesis tasks regarding images~\cite{rombach2022high,zhang2023adding}, videos~\cite{ho2022video,ho2022imagen}, and motions~\cite{MDM,zhang2022motiondiffuse}. To further induce generation results to satisfy specific demands or constraints, injecting additional knowledge (e.g., expected image styles, human-object contact constraints) into diffusion processes is an emerging methodology in recent studies~\cite{xu2023interdiff,li2023controllable,peng2024hoidifftextdrivensynthesis3d,CG-HOI}. For this purpose, one paradigm is explicitly improving intermediate denoising results in inference steps using linear fusions/imputations with conditions~\cite{lugmayr2022repaint,choi2021ilvr,MDM,karunratanakul2023guided}, optimizations with differential energy functions~\cite{dhariwal2021diffusion,li2023controllable,peng2024hoidifftextdrivensynthesis3d,TeSMo}, or guidance from learnable networks~\cite{xu2023interdiff,yuan2023physdiff,janner2022planning, gao2025eigenactorvariantbodyobjectinteraction,tian2025semantic3dhandobjectinteraction}. Another paradigm is to design additional diffusion branches with novel conditions~\cite{ho2022classifier,zhong2025smoodi,xu2024bayesian} or representations~\cite{CG-HOI} that comprise the knowledge, encompassing the challenge of fusing them with existing branches. Classifier-free guidance~\cite{ho2022classifier,zhong2025smoodi} uses a linear combination of results from different branches in inference steps, Xu \textit{et al.}~\cite{xu2024bayesian} propose blending and learnable merging strategies, while CG-HOI~\cite{CG-HOI} presents additional cross-attention modules. Our method follows the second paradigm with relative representation designs and transforms the problem into a graphical model, optimizing it with an explicit synchronization strategy.

\subsection{Motion Synchronization}


Motion Synchronization aims at coordinating motions of different bodies in a multi-body system to satisfy specific inter-body constraints. As technical bases, for specific parameterized closed-form constraints, the solutions in Euler angles, axis-angle, quaternions, and rotation matrices are derived in traditional approaches~\cite{chen2019distributedadaptiveattitudecontrol, jin2020eventtriggeredattitudeconsensus, meng2010distributedfinitetimeattitudecontainment, chaturvedi2011rigidbodyattitudecontrol, sarlette2007autonomousrigidbodysynchronization, igarashi2009passivitybasedattitudesynchronization, zou2019velocityfreecooperativeattitudetracking, arrigoni2016spectralsynchronization}. To achieve global attitude synchronization under mechanical constraints in $SE(3)$, several works adopt gradient flow~\cite{markdahl2022synchronizationriemannianmanifoldsmultiply, markdahl2022highdimensionalkuramotomodelsstiefel}, lifting method~\cite{thunberg2018liftingmethodanalyzingdistributed}, and matrix decomposition~\cite{thunberg2018dynamiccontrollerscolumnsynchronization} techniques. In the trend of graph-based multi-body modeling, existing approaches present coordination control based on general undirected~\cite{ren2009distributedleaderlessconsensus} and directed~\cite{wang2013flockingnetworkeduncertaineulerlagrange, liu2012controlledsynchronizationheterogeneousmanipulators, wang2014consensusnetworkedmechanicalsystemsdelays} graphs, while some works~\cite{hatanaka2012passivitybasedposesynchronization,thunberg2015consensusformationcontrol} further extend to nonlinear configurations and dynamic topologies. These works do not adopt learning-based methods; by contrast, our method uses a generative model to enhance consistency, formulating data-driven motion synchronization. 

\subsection{Modeling Motions in Frequency Domain}
\label{sec:2.3}


Modeling motions in the frequency domain is a widely used strategy in diverse research areas. Animation approaches~\cite{shinya1992stochastic, stam1996multiscale, stam1997stochastic} leverage frequency domain characterization to generate periodic vibrations like fluttering of clouds, smoke, or leaves. Physical simulation methods~\cite{davis2015image, davis2016visual, diener2009wind} simulate physical motions by analyzing the underlying motion dynamics in the frequency domain. Adopting learning-based methods, GID~\cite{li2024generativeimagedynamics} uses diffusion models to capture object motions in the frequency domain, generating complex scenes with realistic periodic motions.
\section{Method}

\label{sec:3}

In this section, we first formulate the problem (Section ~\ref{sec:3.1}), and describe the motion representation we use (Section ~\ref{sec:3.2}). Then we introduce our algorithm of frequency decomposition (Section ~\ref{sec:3.3}) and model architecture (Section ~\ref{sec:3.4}) in detail. Finally, in Section ~\ref{sec:3.5} and ~\ref{sec:3.6}, our two synchronization mechanisms will be elaborated.

\begin{figure*}[h!]
    \centering
    \includegraphics[width=1.0\linewidth]{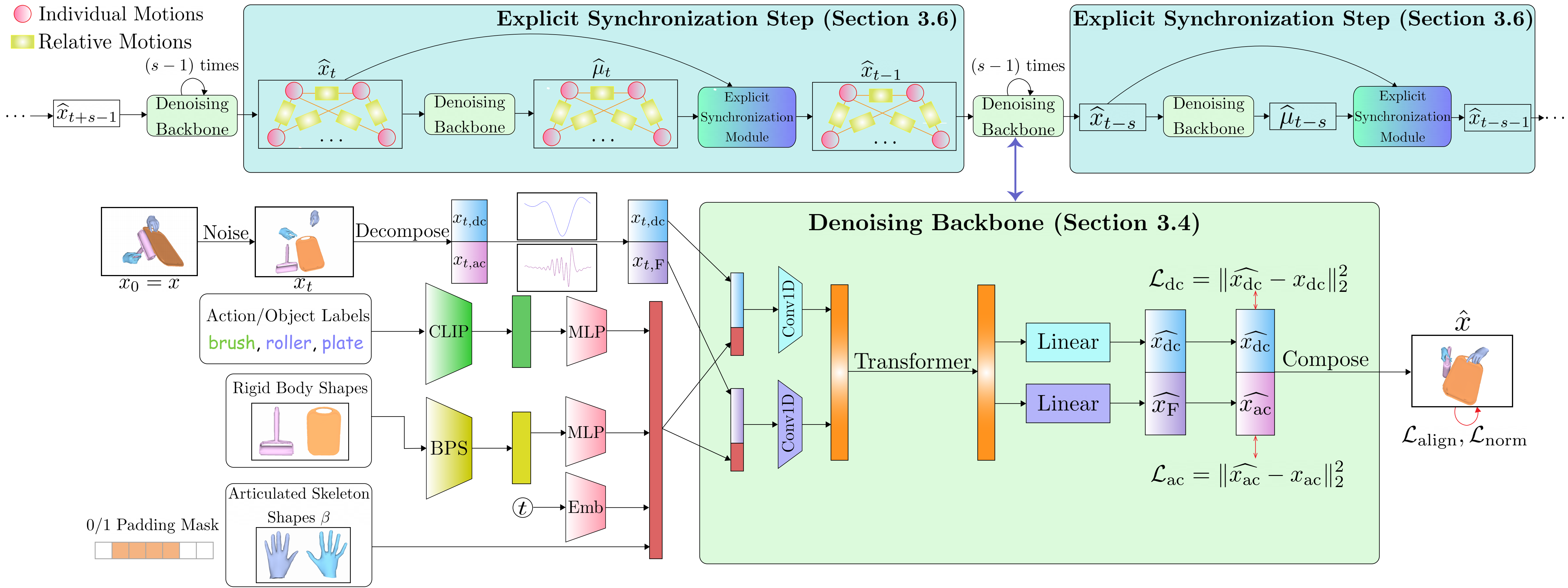}
    \vspace{-0.5cm}
    \caption{\textbf{Overview of SyncDiff.} The light blue boxes show the inference process with explicit synchronization steps performed \textbf{every} $s$ step. For denoising steps irrelevant to explicit synchronization (those marked as ``$(s-1)$ times"), the noise level is set to $\sigma_t$. For the calculation from $\hat{x}$ to $\hat{\mu}$, no noise is added. For the calculation of $\hat{x}_{t-1}$ based on $\hat{x}_t$ and $\hat{\mu}_t$ in explicit synchronization step, the noise level is $\sigma_t^{'}$. Please refer to Section \ref{sec:3.6} for more details. The light green box illustrates the architecture of denoising backbone (Section \ref{sec:3.4}).}
    \vspace{-0.2cm}
    \label{fig:architecture}
\end{figure*}

\subsection{Problem Definition}
\label{sec:3.1}
Consider an interaction scenario comprising $n$ articulated skeletons $h_1, h_2, \dots, h_n$ and $m$ rigid bodies $o_1, o_2, \dots, o_m$, which we uniformly define as ``body". In our experiments, these skeletons are MANO~\cite{Romero_2017} hands or SMPL-X~\cite{pavlakos2019expressivebodycapture3d} humans, whose motions can be reconstructed with joint information and intrinsic shape parameter $\beta$. Details can be found in Section~\ref{sec:B.2}. Our task is to synthesize body motions with known action labels, object categories and geometries, and shape parameters $\beta_{i\in[1,n]}$ of each skeleton. We highlight that our task releases the needs of either predefined object motions \cite{li2023objectmotionguidedhuman} or HOI key frames \cite{christen2024diffh2odiffusionbasedsynthesishandobject} that are used in other existing works.

\subsection{Motion Representation}
\label{sec:3.2}
As mentioned in Section~\ref{sec:1}, in the graphical model, we need individual motions for single bodies on nodes, and relative motions for pairwise bodies on edges.

For a single body $\textbf{b}=h_{i\in [1, n]}$ or $\textbf{b}=o_{j\in[1,m]}$, we represent its motion as $x_{\textbf{b}}$ in the world coordinate system. Here $x_{h_i}\in \mathbb{R}^{N\times 3D }$ represents 3D joint positions, where $N$ is number of frames, and $D$ is number of joints. $x_{o_j}=[\textbf{t}_{o_j}, \textbf{q}_{o_j}]\in \mathbb{R}^{N\times 7}$, where $\textbf{t}_{o_j}\in \mathbb{R}^{N\times 3}$ is its translation, and $\textbf{q}_{o_j}\in \mathbb{R}^{N\times 4}$ is the quaternion for its orientation. 

For two bodies $\textbf{b}_1$ and $\textbf{b}_2$, we use $x_{\textbf{b}_2\rightarrow \textbf{b}_1}$ to denote the relative motion of $\textbf{b}_2$ in $\textbf{b}_1$'s coordinate system. We require that $\textbf{b}_1$ is a rigid body, and omit relative representation between articulated skeletons. If $\textbf{b}_2$ is also a rigid body, then $x_{\textbf{b}_2\rightarrow \textbf{b}_1}=[\textbf{t}_{\textbf{b}_2\rightarrow \textbf{b}_1}, \textbf{q}_{\textbf{b}_2\rightarrow \textbf{b}_1}]$, where $\textbf{t}_{\textbf{b}_2\rightarrow \textbf{b}_1}=\textbf{q}_{\textbf{b}_1}^{-1}(\textbf{t}_{\textbf{b}_2}-\textbf{t}_{\textbf{b}_1})$, and $\textbf{q}_{\textbf{b}_2\rightarrow \textbf{b}_1}=\textbf{q}_{\textbf{b}_1}^{-1}\textbf{q}_{\textbf{b}_2}$.
If $\textbf{b}_2$ is an articulated skeleton, the position of one of its joints $\textbf{p}$ under individual representation will become $\textbf{p}^{'}$ in $\textbf{b}_1$'s coordinate system, where $\textbf{p}^{'}=\textbf{q}_{\textbf{b}_1}^{-1}(\textbf{p}-\textbf{t}_{\textbf{b}_1})$.

After deriving all individual and relative motion representations, we concatenate all of them together into a high-order representation $x$, including $\{x_{o_{j\in[1, m]}}\}$, $\{x_{h_{i\in[1, n]}}\}$, $\{x_{o_{j_2}\rightarrow o_{j_1}}\mid j_1, j_2\in [1, m], j_1\neq j_2\}$, $\{x_{h_i\rightarrow o_j}\mid i\in [1, n], j\in [1, m]\}$. Thus, $x\in \mathbb{R}^{N\times D_{\text{sum}}}$, where $D_{\text{sum}}=7m+3Dn+7m(m-1)+3Dmn$. Although $x$ is composed of individual and relative representations, the final synthesized motions only involve individual motions, and those relative ones serve merely as auxiliary components.

\subsection{Frequency Decomposition}
\label{sec:3.3}

As body number increases, action semantics become more dependent on subtle, high-frequency interactions between bodies (e.g., periodic frictions between the teapot and brush during the action of \emph{brushing}). Unfortunately, using existing works to denoise based on time-domain trajectories often results in these high-frequency components being overshadowed by common low-frequency, large-scale patterns (e.g., objects only move and contact each other, without relative interactions). To better model these high-frequency components, inspired by GID~\cite{li2024generativeimagedynamics}, we explicitly represent them in the frequency domain and supervise them independently in the loss terms described in Section~\ref{sec:3.5}.

For a motion sequence $x \in \mathbb{R}^{N\times D_{\text{sum}}}$, we select $N$ frequency bases $\phi_{l\in[0,N-1]}$=$\frac{l}{N}$, and then decompose $x$ into components of $\phi$ by Fast Fourier Transform (FFT):
\begin{equation}
x_{u\in[0,N-1]}=\sum\limits_{l=0}^{N-1}a_l\cos(u\phi_l)+b_l \sin(u\phi_l),
\end{equation}
where $a_l, b_l\in\mathbb{R}^{D_{\text{sum}}}$ are the coefficients computed by FFT. Note that here $u$ denotes the frame id, which is different from the noise timestep $t$ for the diffusion model later. To prevent networks from overfitting high-frequency noises in mocap datasets, we select a cutoff boundary $L\in[4, \frac{N}{4})$, and directly discard signals with frequencies higher than $\phi_L/2\pi$. In practice, we select $L=16$, which strikes a balance between motion fidelity conservation and simplicity (See Section~\ref{sec:B.1}). We then divide the remaining signals into low-frequency components ($x_{\text{dc}}$) and high-frequency components ($x_{\text{ac}}$):
\begin{equation}
\begin{split}
& x_{\text{dc}, u}=\sum\limits_{l=-3}^2 a_l\cos(u\phi_l)+b_l \sin(u\phi_l), \text{ and} \\
& x_{\text{ac}, u}=\sum\limits_{l\in [-L, -4]\cup[3, L-1]} a_l\cos(u\phi_l)+b_l \sin(u\phi_l),
\end{split}
\end{equation}
where $a_l$=$a_{l+N}$ and $b_l$=$b_{l+N}$ for $l<0$. $x_{\text{dc}}, x_{\text{ac}}\in \mathbb{R}^{N\times D_{\text{sum}}}$. Additionally, we denote the frequency domain representation of high-frequency components as $x_{\text{F}}=[a_3, \dots, a_{L-1}, a_{N-L}, \dots, a_{N-4},  b_3, \dots, b_{L-1}, b_{N-L}, \dots, \\b_{N-4}, z] \in \mathbb{R}^{N\times D_{\text{sum}}}$, where $z$ is a zero mask, padding $x_{\text{F}}$ into the same length as $x_{\text{dc}}$. Our model denoises on $x_{\text{dc}}$ and $x_{\text{F}}$, and supervises the differences on $x_{\text{dc}}$ and $x_{\text{ac}}$ between synthesized and ground-truth values in temporal domain.

\subsection{Model Architecture}
\label{sec:3.4}


We jointly model all individual and relative motions in a high-order representation with one single diffusion model (Figure \ref{fig:architecture}). The latent diffusion~\cite{rombach2022high} paradigm is adopted, where action and object label features are extracted from pretrained CLIP ~\cite{radford2021learningtransferablevisualmodels}, and object geometry features are encoded by BPS ~\cite{prokudin2019efficientlearningpointclouds}. To facilitate batch operations, we pad all trajectories into the same length, and use 0/1 padding masks to indicate the positions that need to be generated. Concatenate together label/geometry features, padding masks, noise timestep embeddings, and shape parameters $\beta_{i\in [1,n]}$ to form the condition vector. The condition vector is replicated and combined with $x_{t, \text{dc}}$ and $x_{t, \text{F}}$ decomposed from noised $x_t$, respectively. They are then projected into the latent space to be denoised by a transformer-based backbone. Let the predicted results be $\widehat{x_{\text{dc}}}$ and $\widehat{x_{\text{F}}}$, the latter of which is reconstructed as $\widehat{x_{\text{ac}}}$. The final denoised motion sequence is recomposed by $\widehat{x}=\widehat{x_{\text{dc}}}+\widehat{x_{\text{ac}}}$. Please refer to Section~\ref{sec:D.2},~\ref{sec:D.3} for model hyperparameters.


\subsection{Loss Functions}
\label{sec:3.5}

To enhance synchronization, besides data scores, we design a set of alignment scores. The idea is, for every edge on the graphical model, the alignment score guides the computed relative motion between two individual motions to approach the generated relative motion, thereby achieving synchronization across the entire graphical model. We further mathematically derive the corresponding loss term $\mathcal{L}_{\text{align}}$.

Suppose we need to supervise the final synthesis results $\{\widehat{x_{\text{dc}}}, \widehat{x_{\text{ac}}}\}$, where $\{x_{\text{dc}}, x_{\text{ac}}\}$ are ground-truth motions. We need loss functions for both data and alignment scores.

For data sample scores, our method is similar to the standard reconstruction loss, except that we supervise $x_{\text{dc}}$ and $x_{\text{ac}}$ separately. We denote $\mathcal{L}_{\text{dc}}=\Vert \widehat{x_{\text{dc}}}-x_{\text{dc}}\Vert_2^2$, $\mathcal{L}_{\text{ac}}=\Vert \widehat{x_{\text{ac}}}-x_{\text{ac}}\Vert_2^2$ and $\mathcal{L}_{\text{norm}}=\sum\limits_{j=1}^m \Vert 1-|\hat{\textbf{q}}_j|\Vert _2^2$. The last one is used to induce the norms of the quaternions representing rigid body rotations to be as close to $1$, where $\hat{x}_{o_j}=[\hat{\textbf{t}}_j, \hat{\textbf{q}}_j]$.      

Define $\text{rel}(x_{\textbf{a}}, x_{\textbf{b}})$ as $\textbf{b}$'s motion relative to $\textbf{a}$. Detailed formulas can be found in Section ~\ref{sec:C.1}. For our alignment scores of pairwise bodies, we can derive the corresponding alignment loss term
\begin{equation}
\begin{aligned}
\mathcal{L}_{\text{align}}= & \sum\limits_{j_1, j_2\in [1, m], j_1\neq j_2}\Vert \hat{x}_{o_{j_2}\rightarrow o_{j_1}}-\text{rel}(\hat{x}_{o_{j_1}}, \hat{x}_{o_{j_2}})\Vert_2^2 \\
& +\sum\limits_{i\in [1,n], j\in [1,m]}\Vert \hat{x}_{h_i\rightarrow o_j}-\text{rel}(\hat{x}_{o_j}, \hat{x}_{h_i})\Vert_2^2,
\end{aligned}
\end{equation}

Finally, the total loss function is calculated as:
\begin{equation}
\mathcal{L}=\lambda_{\text{dc}}\mathcal{L}_{\text{dc}}+\lambda_{\text{ac}}\mathcal{L}_{\text{ac}}+\lambda_{\text{align}}\mathcal{L}_{\text{align}}+\lambda_{\text{norm}}\mathcal{L}_{\text{norm}}.
\end{equation}

Proof details of the equivalence between $\mathcal{L}_{\text{align}}$ and alignment scores are provided in Section~\ref{sec:A.2}.

\subsection{Explicit Synchronization in Inference Time}
\label{sec:3.6}
Relying solely on alignment losses only indirectly enhances synchronization. To directly improve synthesis quality and diversity, as well as provide stronger \textbf{theoretical guarantees}, we introduce an explicit synchronization process, which is mathematically equivalent to maximum total likelihood sampling during inference time, aiming to leverage both data sample scores and alignment scores to address this problem. Since the synchronization step is time-consuming, to balance performance and efficiency, we perform synchronization operations \textbf{every} $s(s\ll T)$ steps, where $T$ is the total number of denoising steps, as is shown in Figure \ref{fig:architecture}. In practice, we take $s=50$, $T=1000$,  to ensure synchronization while improving inference speed, as further detailed in Section~\ref{sec:B.2}. For the predicted motion $\hat{x}_t$ at step $t\in [1, T]$, according to the diffusion formula~\cite{ho2020denoisingdiffusionprobabilisticmodels}, without synchronization, the next step would be:

\begin{equation}
    \hat{x}_{t-1}=\hat{\mu}(\hat{x}_t, t)+\sigma_t \epsilon\quad (\epsilon\sim \mathcal{N}(0, I)),
    \label{eq:sample}
\end{equation}
where $\hat{\mu}$ is the predicted mean value, and noise scale $\sigma_t$ is a predefined constant real number. 
For convenience, we denote the motion before synchronization as $\hat{x}$ and the motion after synchronization as $\hat{x}^{'}$. Let $\sigma_t$ be abbreviated as $\sigma$. 
For different parts of $\hat{x}^{'}$, we handle them as follows:

1. \textbf{Individual Motions of Rigid Bodies}. Let
\begin{equation}
\label{eq:individual_motions_of_rigid_bodies}
\begin{split}
\hat{x}_{o_j}^{'}= & \frac{\frac{2}{m-1}\sigma^2\overline{\lambda}}{1+2\sigma^2\overline{\lambda}}\sum\limits_{j^{'}\neq j}\text{comb}\left(\hat{x}_{o_{j^{'}}}, \hat{x}_{o_j\rightarrow o_{j^{'}}}\right) \\
& +\frac{1}{1+2\sigma^2\overline{\lambda}}\hat{\mu}_{o_j}+\sigma^{'}\epsilon,
\end{split}
\end{equation}
where $\text{comb}(x_{\textbf{a}}, x_{\textbf{b}\rightarrow \textbf{a}})$ utilizes the individual motion of $\textbf{a}$ and relative motion between $\textbf{b}$ and $\textbf{a}$ to compute $\textbf{b}$'s motion. Its formula is in Section~\ref{sec:C.1}. $\overline{\lambda}=\frac{\lambda_{\text{exp}}}{R}\sum\limits_{r=1}^R \frac{1}{2\sigma_{t_r}^2}$, where $R=T/s=20$ is the number of synchronization steps, $t_1, t_2, \dots, t_R$ are the corresponding timesteps, and $\sigma_{t_1}$, $\sigma_{t_2}$, $\dots$, $\sigma_{t_R}$ are the original correspondent noise scales (without synchronization). Finally, $\sigma^{'}=\sqrt{\frac{\sigma^2}{1+2\sigma^2\overline{\lambda}}}$. Specifically, when $m=1$, we do not perform synchronization for this part, and the denoising formula is identical to that without synchronization.
    
2. \textbf{Individual Motions of Articulated Skeletons}. Let
\begin{equation}
\label{eq:individual_motions_of_articulated_skeletons}
\begin{split}
\hat{x}_{h_i}^{'}= & \frac{\frac{2}{m}\sigma^2\overline{\lambda}}{1+2\sigma^2\overline{\lambda}}\sum\limits_{j\in [1, m]}\text{comb}\left(\hat{x}_{o_j}, \hat{x}_{h_i\rightarrow o_j}\right) \\
& +\frac{1}{1+2\sigma^2\overline{\lambda}}\hat{\mu}_{h_i}+\sigma^{'}\epsilon,
\end{split}
\end{equation}

Definitions of $\overline{\lambda}$ and $\sigma^{'}$ are the same as Eq.~\ref{eq:individual_motions_of_rigid_bodies}.
    
3. \textbf{Relative Motions}. Let
\begin{equation}
\label{eq:individual_motions_of_relative}
\begin{split}
&\hat{x}_{{o_j\rightarrow o_{j^{'}}}}^{'}=\frac{1}{1+2\sigma^2\overline{\lambda}}\hat{\mu}_{{o_j\rightarrow o_{j^{'}}}}+\frac{2\sigma^2\overline{\lambda}}{1+2\sigma^2\overline{\lambda}}\text{rel}\left(\hat{x}_{o_{j^{'}}}, \hat{x}_{o_j}\right)+\sigma^{'}\epsilon, \\
&\hat{x}_{h_i\rightarrow o_j}^{'}=\frac{1}{1+2\sigma^2\overline{\lambda}}\hat{\mu}_{h_i\rightarrow o_j}+\frac{2\sigma^2\overline{\lambda}}{1+2\sigma^2\overline{\lambda}}\text{rel}\left(\hat{x}_{o_j}, \hat{x}_{h_i}\right)+\sigma^{'}\epsilon.
\end{split}
\end{equation}

Definitions of $\overline{\lambda}$ and $\sigma^{'}$ are the same as Eq.~\ref{eq:individual_motions_of_rigid_bodies}. Function rel is defined in Section~\ref{sec:3.5}.

After synchronization operations and deriving $\hat{x}^{'}$, it is again used for further denoising. Between two adjacent synchronization steps, we still directly use Eq. \ref{eq:sample} for stepwise denoising, and do not perform synchronization operations. We demonstrate in Section~\ref{sec:A.2}, that the above equations are equivalent to maximum likelihood sampling from the newly computed Gaussian distribution based on both data sample scores and alignment scores.

\section{Experiments}
\label{sec:4}
In this section, we first introduce some basic experimental settings (Sections~\ref{sec:4.1},~\ref{sec:4.2}), and then demonstrate the visual and quantitative results in comparison to several baselines (Section~\ref{sec:4.3}) and ablation settings (Section~\ref{sec:4.4}). We also provide in-depth analyses for the results.
\subsection{Datasets}
\label{sec:4.1}

To examine our method's generalizability across various multi-body interaction configurations, we utilize five datasets with different interaction scenarios: TACO~\cite{liu2024tacobenchmarkinggeneralizablebimanual} (two hands and two objects), CORE4D~\cite{zhang2024core4d4dhumanobjecthumaninteraction} (two people and one object), GRAB~\cite{taheri2020grab} (one or two hands and one object), OAKINK2~\cite{zhan2024oakink2datasetbimanualhandsobject} (two hands and one to three objects), and BEHAVE~\cite{bhatnagar2022behave} (one human and one object). We describe data splits for each dataset below. Detailed dataset statistics and split sizes are shown in Section~\ref{sec:D.1}.

(1) \textbf{TACO}: We use the official split of TACO, with four testing sets, each representing different scenarios: 1) the interaction triplet $\langle$action, tool category, target object category$\rangle$ and the object geometries are all seen in the training set, 2) unseen object geometry, 3) unseen triplet, and 4) unseen object category and geometry. 

(2) \textbf{CORE4D}: We divide $875$ motion sequences into one training set and two testing sets, where the two testing sets represent seen and unseen object geometries, respectively. The $\langle$action, object category$\rangle$ pairs from testing sets are all involved in the training set. 

(3) \textbf{GRAB}: We use an existing data split of unseen subjects from IMoS~\cite{IMoS} and that of unseen objects from DiffH$_{2}$O~\cite{christen2024diffh2odiffusionbasedsynthesishandobject}. Please refer to the two papers for details.

(4) \textbf{OAKINK2}: We utilize the train, val, and test divisions stated in the TaMF task of their paper.

(5) \textbf{BEHAVE}: We utilize the official splits of train/test.

Toward a fair comparison with DiffH$_{2}$O~\cite{christen2024diffh2odiffusionbasedsynthesishandobject}, we follow the setting of DiffH$_{2}$O for GRAB, where methods are required to generate hand-object manipulations after grasping objects.
For other four datasets, all methods need to synthesize complete multi-body interaction motion sequences.

\subsection{Evaluation Metrics}
\label{sec:4.2}

To evaluate the qualities of synthesized motion sequences comprehensively, we present two types of evaluation metrics focusing on fine-grained contact consistency and general motion semantics, respectively.

(1) \textbf{Contact-based metrics} measure the contact plausibility of hand-object/human-object interactions and the extent of motion coordination among different bodies. We use the Contact Surface Ratio (CSR) for hand-object settings and the Contact Root Ratio (CRR) for human-object settings to denote the proportion of frames where hand-object/human-object contact occurs. \textbf{Contact} is defined as the hand mesh being within $5$mm of at least one object for CSR, and the two hand roots of a human consistently being within $3$cm of at least one object for CRR. When there are multiple hands or humans, we take the average among all of them. We label the frames based on whether hand-object contact occurs for ground-truth or synthesized motions, and then compute their Intersection-over-Union (IoU), denoted as CSIoU. To eliminate potential ambiguities, we've provided corresponding pseudocodes in Section~\ref{sec:C.5}. Besides, Interpenetration Volume (IV) and Interpenetration Depth (ID) are incorporated to assess penetration between different bodies.

(2) \textbf{Motion semantics metrics} evaluate high-level motion semantics and its distributions, comprising Recognition Accuracy (RA), Fréchet Inception Distance (FID), Sample Diversity (SD), and Overall Diversity (OD).
Following existing evaluations for motion synthesis~\cite{li2023objectmotionguidedhuman,li2023controllable}, we train a network to extract motion features and predict action labels using ground-truth motion data. For better feature semantics, the network is trained on the combination of all train, val, and test splits. RA denotes the action recognition accuracy of the network on synthesized motions. FID measures the difference in feature distributions of generated and ground-truth motions. Following DiffH$_{2}$O~\cite{christen2024diffh2odiffusionbasedsynthesishandobject}, SD represents the mean Euclidean distance between multiple generated wrist trajectories in a single sample, and OD refers to mean distance between all generated trajectories in a dataset split. To measure semantics from a perceptual aspect, we conduct user studies, comparing our methods with baselines, which is shown in Section~\ref{sec:4.5}.

\subsection{Comparison to Existing Methods}
\label{sec:4.3}

\begin{table*}[t!]
\centering
\small
\addtolength{\tabcolsep}{-3pt}
{
\begin{tabular}{c|c c c c|c c c c|c c c c|c c c c}
\toprule
\multirow{2}{*}{Method}&  \multicolumn{4}{c|}{CSIoU (\%, $\uparrow$)} & \multicolumn{4}{c|}{IV ($\text{cm}^3$, $\downarrow$)} & \multicolumn{4}{c|}{FID ($\downarrow$)} & \multicolumn{4}{c}{RA (\%, $\uparrow$)}\\
\cline{2-17}
 & Test1 & Test2 & Test3 & Test4 & Test1 & Test2 & Test3 & Test4 & Test1 & Test2 & Test3 & Test4 & Test1 & Test2 & Test3 & Test4\\
\midrule
Ground-truth & 100.0 & 100.0 & 100.0 & 100.0 & 4.56 & 3.60 & 4.24 & 3.50 & 0.03 & 0.03 & 0.03 & 0.04 & 84.92 & 89.00 & 75.86 & 65.90\\

MACS~\cite{shimada2023macsmassconditioned3d} & 56.81 & 53.79 & 21.38 & 12.09 & 13.18 & 18.57 & 10.33 & 8.99 & 10.56 & 23.24 & 32.18 & 42.37 & 58.40 & 53.08 & 33.00 & 19.02\\

DiffH$_{2}$O~\cite{christen2024diffh2odiffusionbasedsynthesishandobject} & 62.29 & 46.38 & 42.12 & 16.38 & 10.25 & 15.21 & 4.67 & \textbf{5.70} & 4.34 & 17.04 & 24.92 & 39.20 & 61.40 & 56.70 & 43.67 & 28.15\\

\textbf{SyncDiff (Ours)} & \textbf{73.00} & \textbf{70.94} & \textbf{43.22} & 26.70 & 6.64 & \textbf{3.81} & \textbf{4.02} & 7.73 & \textbf{2.70} & \textbf{2.68} & 22.96 & \textbf{30.23} & \textbf{73.28} & \textbf{85.92} & \textbf{46.90} & \textbf{40.12}\\

\midrule
w/o all & 62.96 & 52.38 & 38.02 & 26.39 & 7.95 & 12.02 & 7.05 & 7.67 & 10.63 & 21.87 & 30.17 & 46.38 & 57.39 & 48.05 & 37.13 & 24.92\\
w/o decompose & 68.86 & 54.77 & 41.70 & \textbf{28.07} & 6.80 & 10.78 & 6.93 & 7.22 & 6.44 & 21.21 & 28.67 & 49.58 & 56.60 & 51.85 & 40.02 & 22.18\\
w/o $\mathcal{L}_{\text{align}}$, exp sync & 63.74  & 48.35 & 39.89 & 20.87 & 14.28 & 13.80 & 5.93 & 7.44 & 4.13 & 4.32 & 24.65 & 38.73 & 64.47 & 62.12 & 41.68 & 30.39\\
w/o $\mathcal{L}_{\text{align}}$ & 70.39 & 67.15 & 40.38 & 26.83 & \textbf{6.29} & 4.86 & 5.88 & 7.39 & 2.90 & 3.02 & 22.28 & 32.78 & 67.82 & 79.30 & 44.75 & 34.51\\
w/o exp sync & 65.51 & 50.33 & 37.72 & 23.61 & 13.08 & 14.40 & 6.20 & 7.75 & 3.39 & 3.30 & \textbf{21.26} & 33.67 & 67.27 & 78.50 & 45.82 & 37.13\\
\bottomrule
\end{tabular}
}
\vspace{-0.2cm}
\caption{Results on TACO~\cite{liu2024tacobenchmarkinggeneralizablebimanual} dataset. The best in each column is highlighted in bold.}
\vspace{-0.7cm}
\label{tab:table1}
\end{table*}

\textbf{Hand-Object Interaction.} For hand-object interaction motion synthesis, we compare our method to two state-of-the-art approaches, MACS~\cite{shimada2023macsmassconditioned3d} and DiffH$_{2}$O~\cite{christen2024diffh2odiffusionbasedsynthesishandobject}. For MACS, we first generate the motions of all objects in their object trajectory synthesis phase, and then directly use their hand motion synthesis phase to get overall results. For DiffH$_{2}$O, we use the version without grasp reference input.

\begin{figure}[h!]
    \centering
    \includegraphics[width=1.0\linewidth]{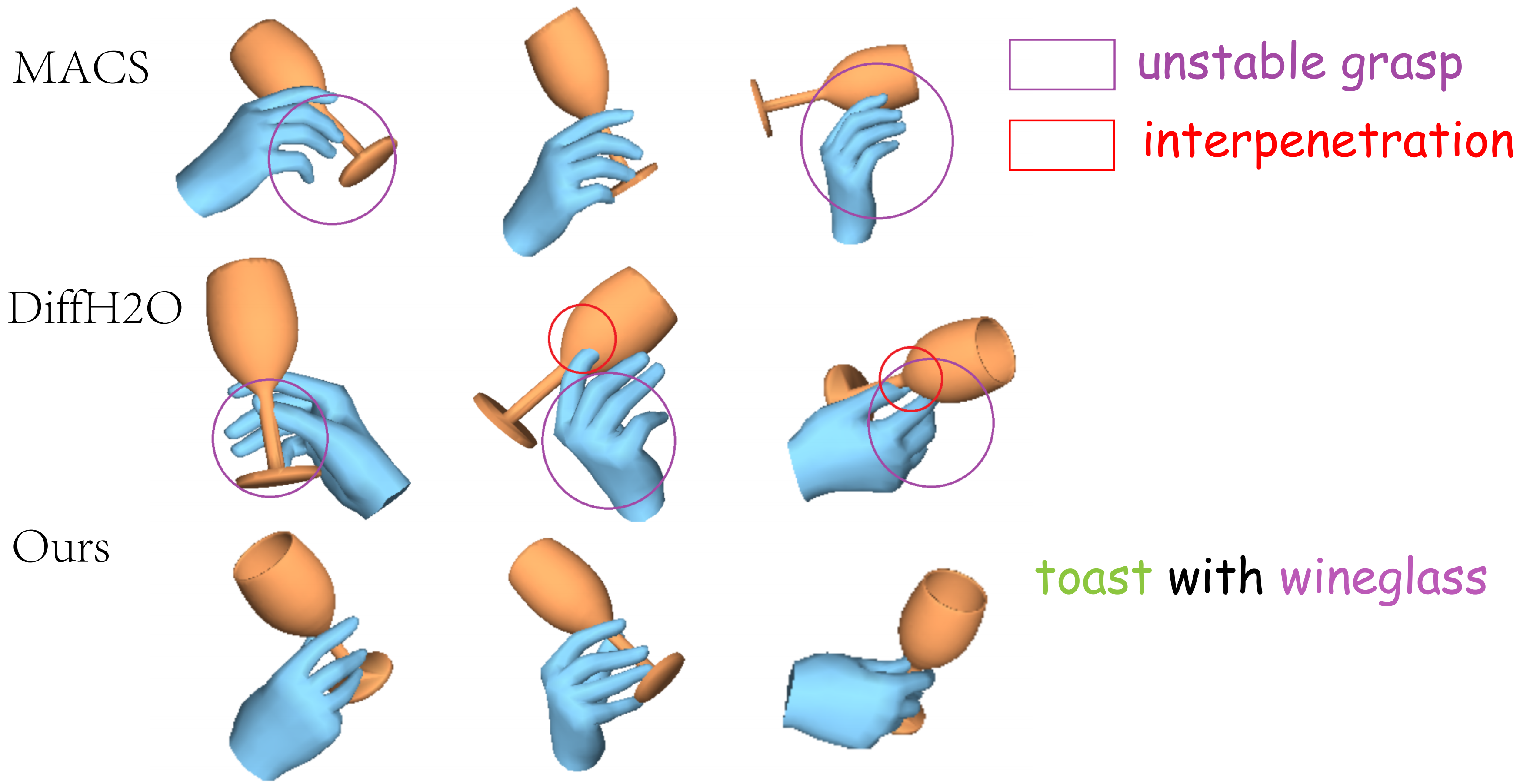}
    \vspace{-0.5cm}
    \caption{Qualitative results from GRAB~\cite{taheri2020grab} dataset.}
    \vspace{-0.4cm}
    \label{fig:GRAB}
\end{figure}

\begin{figure}[h!]
    \centering
    \includegraphics[width=1.0\linewidth]{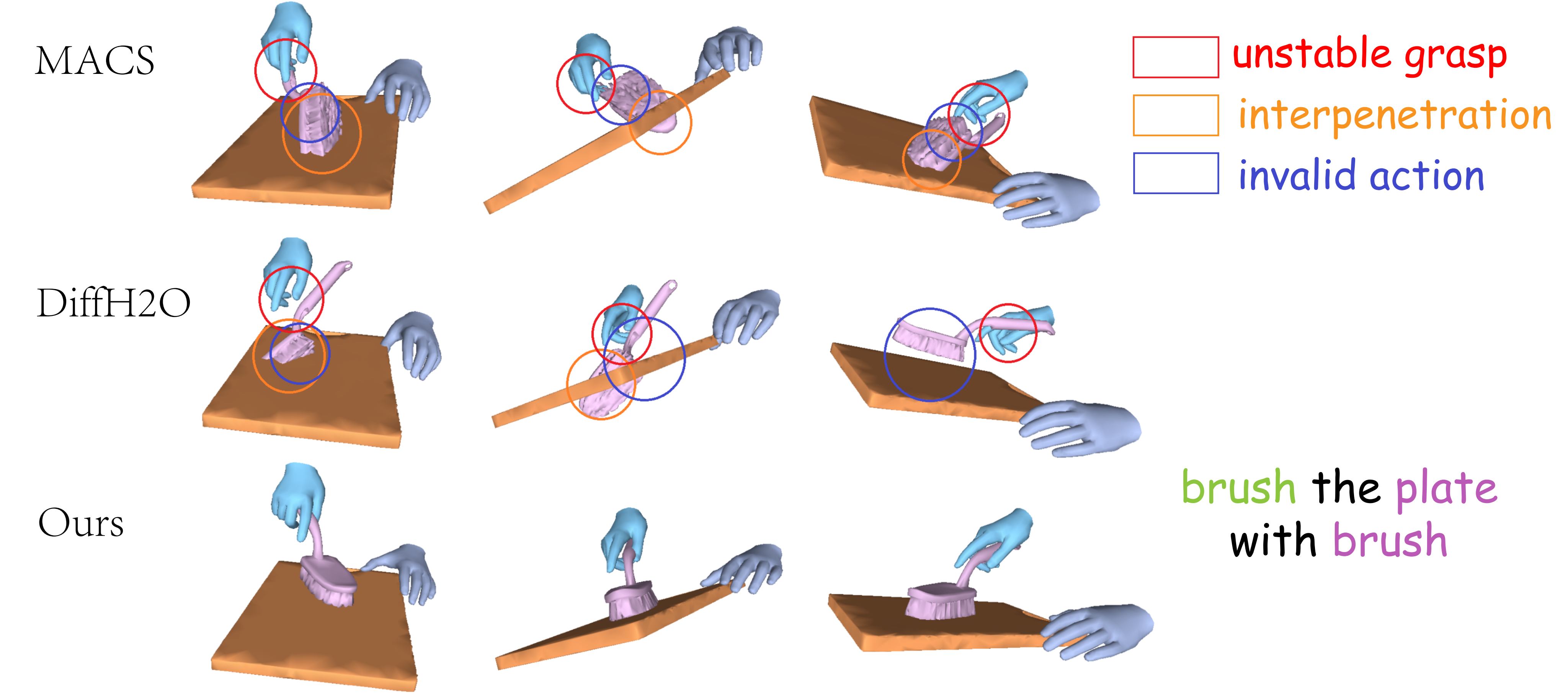}
    \vspace{-0.8cm}
    \caption{Qualitative results from TACO~\cite{liu2024tacobenchmarkinggeneralizablebimanual} dataset. Invalid action indicates the poses cannot complete the operation effectively.}
    \vspace{-0.3cm}
    \label{fig:TACO}
\end{figure}

\begin{figure}[h!]
    \centering
    \vspace{-0.1cm}
    \includegraphics[width=1.0\linewidth]{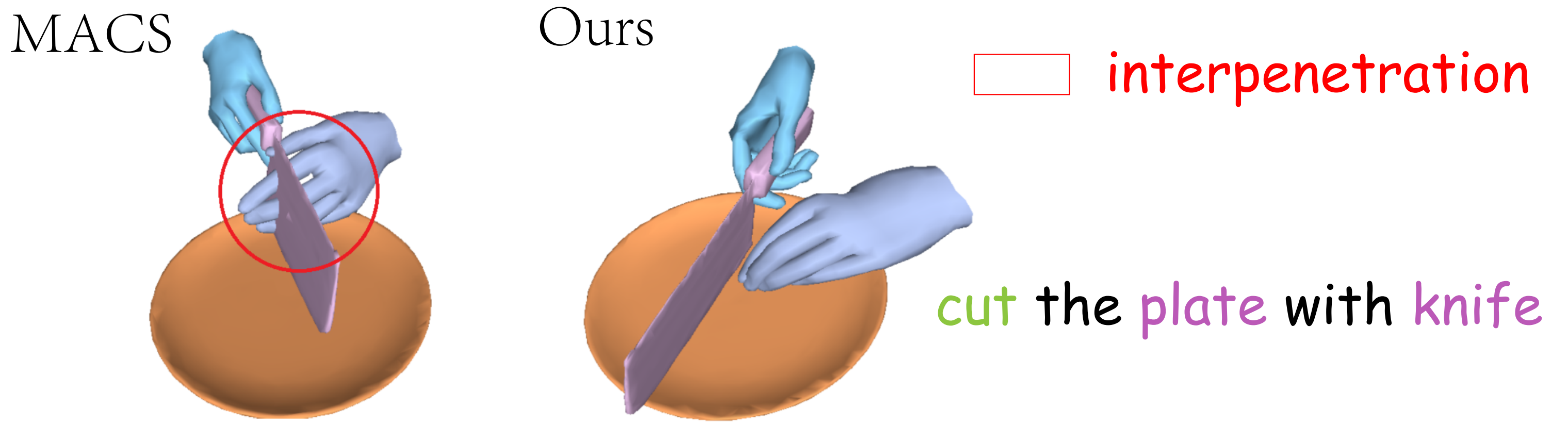}
    \vspace{-0.5cm}
    \caption{Qualitative results from TACO~\cite{liu2024tacobenchmarkinggeneralizablebimanual} dataset.}
    \vspace{-0.4cm}
    \label{fig:TACO3}
\end{figure}

\begin{figure}[h!]
    \centering
    \includegraphics[width=1.0\linewidth]{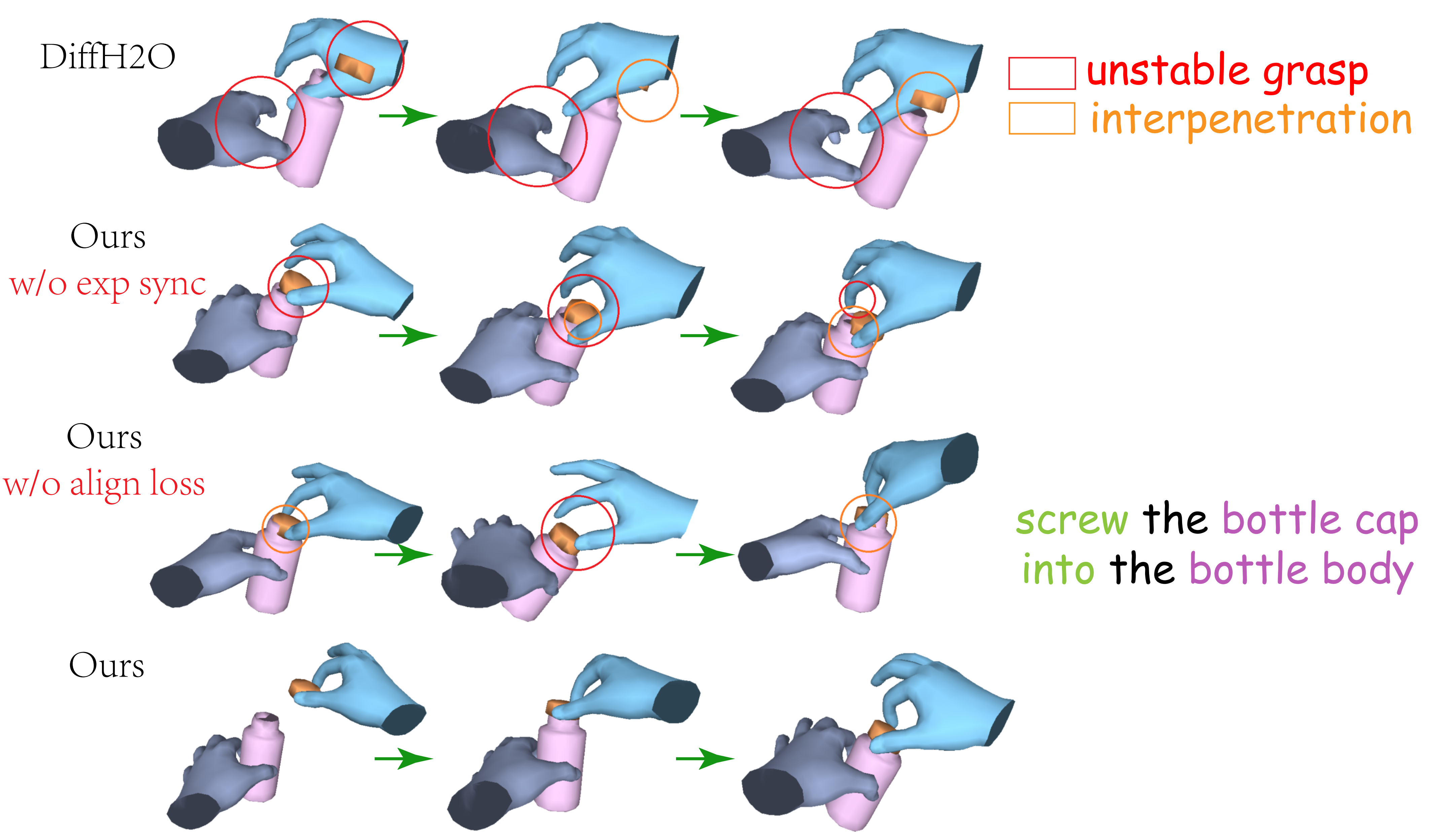}
    \vspace{-0.6cm}
    \caption{Qualitative results from OAKINK2~\cite{zhan2024oakink2datasetbimanualhandsobject} dataset. The task requires precise contact between objects, where the bottle cap needs to align perfectly with the bottle, and there needs to be a tendency for it to be twisted down in a clockwise spiral.}
    \vspace{-0.7cm}
    \label{fig:OAKINK2}
\end{figure}

In Figure~\ref{fig:GRAB}, our method features more realistic contacts and more stable grasping. As is shown in Table \ref{tab:table1}, \ref{tab:table4}, and \ref{tab:table3}, our method outperforms MACS and DiffH$_{2}$O by a large margin with better CSIoU, CSR, IV, and ID. The reason is that our method features more robust alignment and synchronization mechanisms to ensure synchronization which are excluded in existing methods.
In the results from MACS and DiffH$_{2}$O in Figure~\ref{fig:TACO}, the brush often penetrates the plate, and the pose of the brush does not guarantee the bristles being pressed closely against the plate, effectively completing the action. FID and RA further indicate that the motions generated by our method are more semantically realistic. This is caused by the more precise object-object interactions in SyncDiff, and our separation of motions at different frequencies ensures that subtle high-frequency periodic movements are not overshadowed, which are crucial for identifying action types. Figure~\ref{fig:TACO3} demonstrates the benefits of our synchronization mechanism. Although MACS uses relative representations for some hand-object pairs to ensure firm grasp, due to the lack of synchronization on the complete graphical model, conflicts arise between the left hand and the knife, which should not directly interact. Figure \ref{fig:OAKINK2} indicates that with a higher demand for coordination quality among objects, our method can also address hand-object and object-object synchronization effectively. 

\textbf{Human-Object Interaction.} For human-object interaction synthesis, we compare our method to OMOMO~\cite{li2023objectmotionguidedhuman} and CG-HOI~\cite{CG-HOI}. For OMOMO, we first use their conditional diffusion models to generate object trajectories, and then use their whole pipeline to synthesize complete Multi-body HOI motion. Cross-attention in CG-HOI is modified for two humans, one object and contact between them. 

\begin{figure}[h!]
    \centering
    \includegraphics[width=1.0\linewidth]{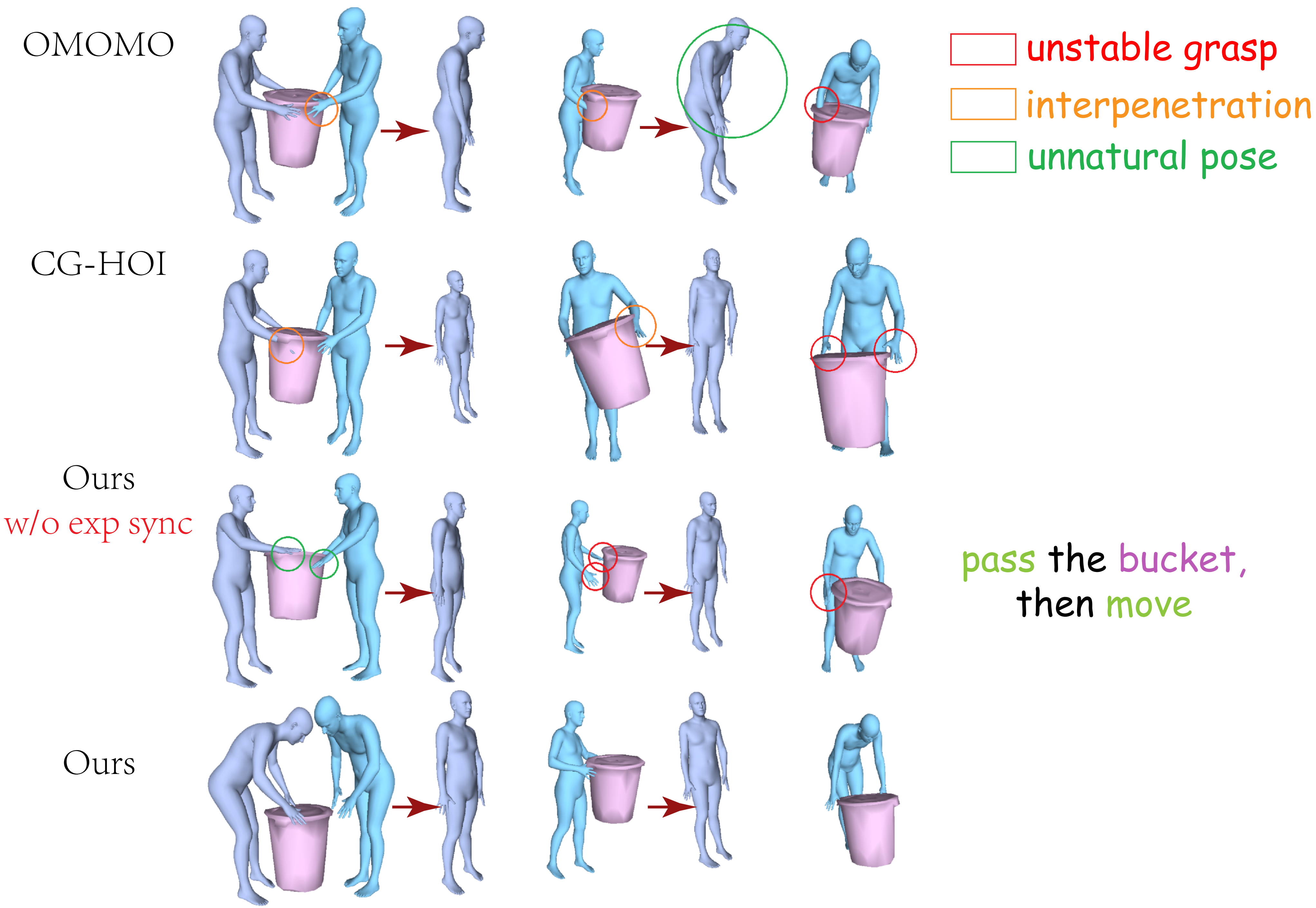}
    \vspace{-0.6cm}
    \caption{Qualitative comparisons on CORE4D~\cite{zhang2024core4d4dhumanobjecthumaninteraction} dataset.}
    \vspace{-0.2cm}
    \label{fig:CORE4D}
\end{figure}

\begin{figure}[h!]
    \centering
    \includegraphics[width=1.0\linewidth]{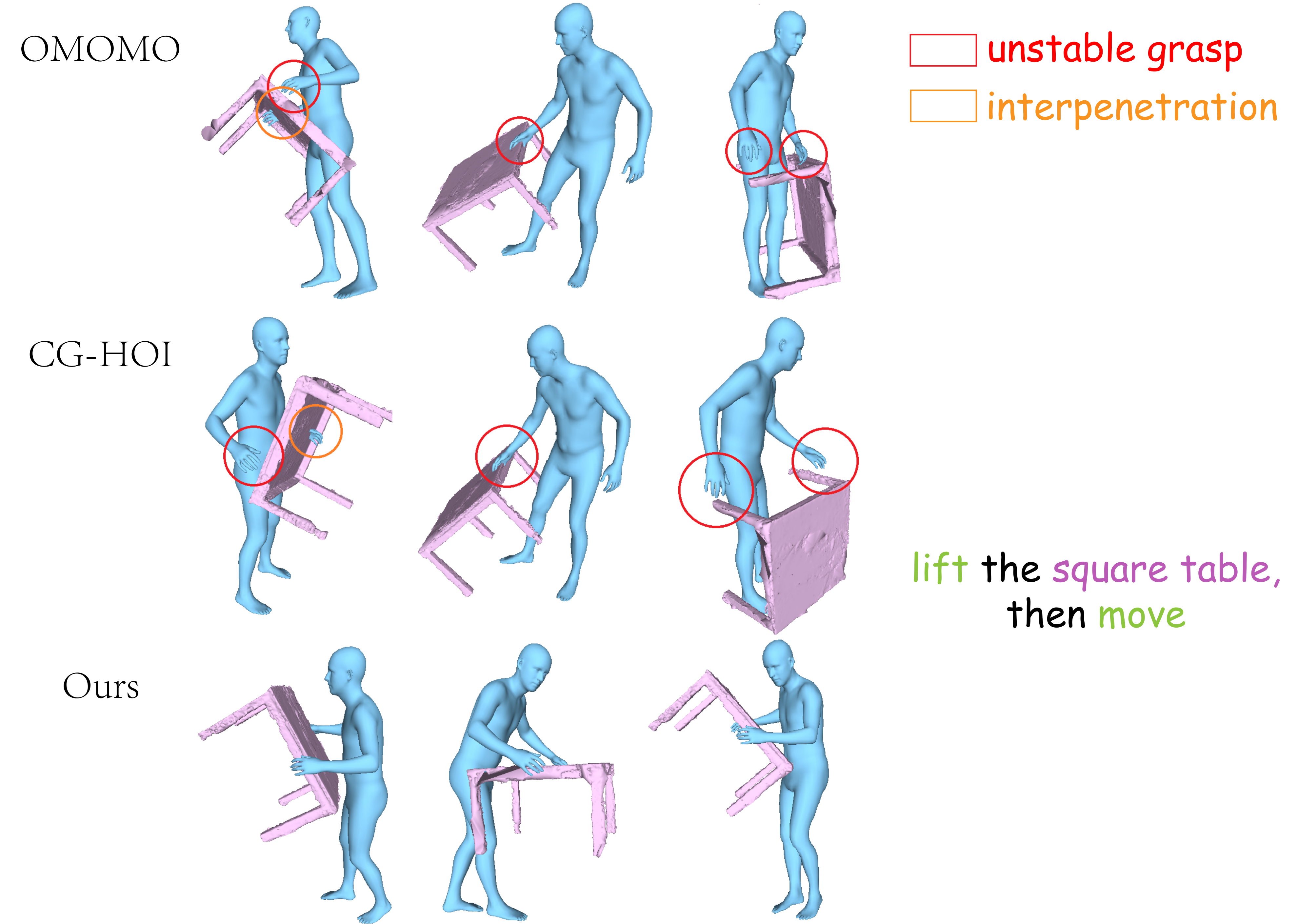}
    \vspace{-0.6cm}
    \caption{Qualitative results from BEHAVE~\cite{bhatnagar2022behave} dataset. Baseline methods suffer from unreasonable grasp poses due to unsynchronized synthesis of body transformations.}
    \vspace{-0.7cm}
    \label{fig:BEHAVE}
\end{figure}

As shown in Tables~\ref{tab:table2},~\ref{tab:table_behave} and Figure~\ref{fig:CORE4D},~\ref{fig:BEHAVE}, our method outperforms CG-HOI and OMOMO in contact-based and semantics metrics and obtains the most realistic qualities. Synthesized motions from existing methods suffer from unnatural grasping poses and arm-object interpenetration, while our method mitigates these issues.

\begin{table*}
  \centering
    \resizebox{1.0\textwidth}{!}{%
  \begin{tabular}{c|c|c|c|c|c|c|c|c|c}
    \toprule
      & Method & Backbone  & SD ($\text{m}, \uparrow$) & OD ($\text{m}, \uparrow$) & IV ($\text{cm}^3, \downarrow$) & ID ($\text{mm}, \downarrow$) & CSR ($\%, \uparrow$) &  \begin{tabular}[c]{@{}c@{}}Hand Motion \\ RA ($\%, \uparrow$)\end{tabular}  & \begin{tabular}[c]{@{}c@{}} Hand-Object Motion\\ RA ($\%, \uparrow$)\end{tabular} \\ 
      \midrule
    \parbox[t]{11mm}{\multirow{5}{*}{\rotatebox[origin=c]{90}{ \begin{tabular}[c]{@{}c@{}} Unseen \\ subject \\ split \end{tabular}}}} 
    & IMoS~\cite{IMoS} & CVAE & 0.002 & 0.149  & 7.14 & 11.47 & 5.0 &  57.9 & 58.8 \\
    & DiffH$_{2}$O~\cite{christen2024diffh2odiffusionbasedsynthesishandobject} & Transformer & 0.088 & 0.185 & 6.65 & 8.39 & \textcolor{blue}{\textbf{6.7}} & 76.0 & 81.0 \\
    & DiffH$_{2}$O~\cite{christen2024diffh2odiffusionbasedsynthesishandobject} & UNet & \textcolor{red}{\textbf{0.109}} & \textcolor{red}{\textbf{0.188}} & \textcolor{red}{\textbf{6.02}} & \textcolor{blue}{\textbf{7.92}} & 6.4 & \textcolor{red}{\textbf{83.3}} & \textcolor{blue}{\textbf{87.5}} \\ 
    & MACS~\cite{shimada2023macsmassconditioned3d}  & MLP+Conv & 0.059 & 0.164  & 8.29 & 10.12 & 3.9 & 72.9 & 76.4 \\
    & \textbf{SyncDiff (Ours)}  & Transformer & \textcolor{blue}{\textbf{0.106}} & \textcolor{red}{\textbf{0.188}}  & \textcolor{blue}{\textbf{6.22}} & \textcolor{red}{\textbf{7.75}} & \textcolor{red}{\textbf{7.2}} & \textcolor{blue}{\textbf{82.6}} & \textcolor{red}{\textbf{88.9}} \\ \midrule
    \parbox[t]{11mm}{\multirow{5}{*}{\rotatebox[origin=c]{90}{\begin{tabular}[c]{@{}c@{}} Unseen \\ object \\ split \end{tabular}}}} 
    & IMoS~\cite{IMoS}  & CVAE & 0.002 & 0.132  & 10.38 & 12.45 & 4.8 & 56.1 & 58.1\\
    & DiffH$_{2}$O~\cite{christen2024diffh2odiffusionbasedsynthesishandobject} & Transformer & 0.133 & \textcolor{blue}{\textbf{0.185}} & \textcolor{blue}{\textbf{7.99}} & \textcolor{blue}{\textbf{10.87}} & 7.3 & 75.0 & 80.3 \\
    & DiffH$_{2}$O~\cite{christen2024diffh2odiffusionbasedsynthesishandobject}  & UNet & \textcolor{blue}{\textbf{0.134}} & 0.179  & 9.03 & 11.39 & \textcolor{blue}{\textbf{8.6}} & \textcolor{blue}{\textbf{75.5}} & \textcolor{blue}{\textbf{83.7}} \\
    & MACS~\cite{shimada2023macsmassconditioned3d}  & MLP+Conv & 0.105 & 0.156  & 11.24 & 13.42 & 5.4 & 57.7 & 63.9 \\
    & \textbf{SyncDiff (Ours)}  & Transformer & \textcolor{red}{\textbf{0.148}} & \textcolor{red}{\textbf{0.192}}  & \textcolor{red}{\textbf{7.07}} & \textcolor{red}{\textbf{10.67}} & \textcolor{red}{\textbf{10.5}} & \textcolor{red}{\textbf{77.4}} & \textcolor{red}{\textbf{86.5}} \\ \bottomrule
  \end{tabular}
  }
  \vspace{-0.2cm}
  \caption{\textbf{Comparison on GRAB~\cite{taheri2020grab} dataset for the post-grasping phase.}
  Following DiffH$_{2}$O~\cite{christen2024diffh2odiffusionbasedsynthesishandobject}, we conduct experiments on the phase where the object has been grasped. Each column highlights the best method in red, with the second best highlighted in blue. Results of IMoS~\cite{IMoS} and DiffH$_{2}$O are from the original paper of DiffH$_{2}$O, while MACS~\cite{shimada2023macsmassconditioned3d} results are obtained via our re-implementation.}
  \vspace{-0.5cm}
  \label{tab:table4}
\end{table*}

\subsection{Ablation studies}
\label{sec:4.4}

We examine the effect of our three key designs (frequency-domain motion decomposition, the alignment loss $\mathcal{L}_{\text{align}}$, and the explicit synchronization) separately. The results after removing each of these three components individually are shown as ``w/o decompose", ``w/o $\mathcal{L}_{\text{align}}$", and ``w/o exp sync". Two additional ablations are to remove the two synchronization mechanisms and all three designs, denoted as ``w/o $\mathcal{L}_{\text{align}}$, exp sync" and ``w/o all", respectively. More details can be found in Section~\ref{sec:C.4}. Results in Tables~\ref{tab:table1}, \ref{tab:table2}, \ref{tab:table3} show that removing any of the three components can lead to varying extents of performance decline. 

\begin{figure}[h!]
    \centering
    \includegraphics[width=1.0\linewidth]{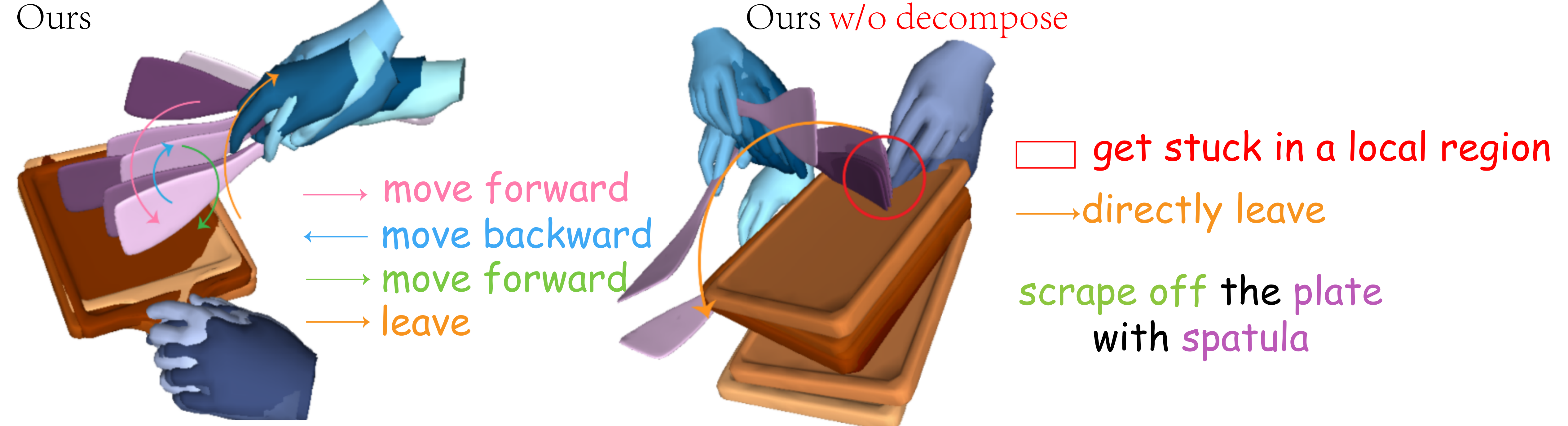}
    \vspace{-0.4cm}
    \caption{Qualitative results from TACO~\cite{liu2024tacobenchmarkinggeneralizablebimanual} dataset. Periodic relative motions are required between two objects. The color changes from deep to light, representing time passage. After removing the decomposition mechanism, the spatula tends to get stuck in a small area on the plate's surface, without effective relative movements.}
    \vspace{-0.3cm}
    \label{fig:TACO2}
\end{figure}

\textbf{Removal of Decomposition.} As shown in Figure~\ref{fig:TACO2}, after removing the decomposition mechanism, once periodic relative motion is involved in the interactions between objects, it becomes easier for high-frequency motions to be neglected, which is intuitively shown as two objects being in an almost relative stationary state, making it difficult to complete the action. This also results in poor performance in semantics metrics FID and RA from Tables~\ref{tab:table1},~\ref{tab:table2}, and~\ref{tab:table3}. In Section~\ref{sec:B.3}, we've provided more detailed experiments to examine the necessity of decomposition in modeling high-frequency components with semantics, through which we have determined that the performance of frequency decomposition significantly surpasses that of merely removing too high-frequency noise.

\begin{table}[t]
\centering
\small
\addtolength{\tabcolsep}{-3pt}
{
\begin{tabular}{c|c c|c c|c c}
\toprule 
\multirow{2}{*}{Method} &  \multicolumn{2}{c|}{CRR(\%, $\uparrow$)} & \multicolumn{2}{c|}{FID($\downarrow$)} & \multicolumn{2}{c}{RA (\%, $\uparrow$)}\\
\cline{2-7}
 & Test1 & Test2 & Test1 & Test2 & Test1 & Test2\\
\midrule
Ground-truth & 7.72 & 6.25  & 0.01 & 0.00 & 96.45 & 97.44\\

OMOMO~\cite{li2023objectmotionguidedhuman} & 5.31 & 5.54 & 13.22 & 14.94 & 68.02 & 65.13\\

CG-HOI~\cite{CG-HOI} & 5.74 & 5.50 & 12.16 & 15.37 & 70.05 & 66.15\\

\textbf{SyncDiff (Ours)} & \textbf{6.15} & \textbf{5.78} & \textbf{6.45} & \textbf{7.25} & \textbf{92.89} & \textbf{90.26}\\

\midrule
w/o all & 5.42 & 5.35 & 17.21 & 21.37 & 54.82 & 48.72\\
w/o decompose & 5.70 & 5.46 & 8.42 & 9.54 & 75.13 & 71.28\\
w/o $\mathcal{L}_{\text{align}}$, exp sync & 4.84 & 4.88 & 7.43 & 8.49 & 82.74 & 74.11\\
w/o $\mathcal{L}_{\text{align}}$ & 5.38 & 5.25 & 6.74 & 8.31 & 90.36 & 87.69\\
w/o exp sync & 5.23 & 5.04 & 7.55 & 7.89 & 80.20 & 78.46\\
\bottomrule
\end{tabular}
}
\vspace{-0.2cm}
\caption{Results on CORE4D~\cite{zhang2024core4d4dhumanobjecthumaninteraction} dataset. The best in each column is highlighted in bold.}
\vspace{-0.3cm}
\label{tab:table2}
\end{table}

\begin{table}[h!]
\centering
\small
\addtolength{\tabcolsep}{-3pt}
{
\begin{tabular}{c c c c}
\toprule
Method & CRR(\%, $\uparrow$) & FID($\downarrow$) & RA (\%, $\uparrow$)\\
\midrule
Ground-truth & 13.02 & 0.03 & 97.78 \\

OMOMO~\cite{li2023objectmotionguidedhuman} & 8.81 & 6.19 &  68.89 \\

CG-HOI~\cite{CG-HOI} & 8.64 & 5.50 & 70.00 \\

\textbf{SyncDiff (Ours)} & \textbf{10.29} & \textbf{4.45} & \textbf{81.11} \\

\bottomrule
\end{tabular}
}
\vspace{-0.2cm}
\caption{Results on BEHAVE~\cite{bhatnagar2022behave} dataset. The best in each column is highlighted in bold.}
\vspace{-0.2cm}
\label{tab:table_behave}
\end{table}

\begin{figure*}[htbp]
    \centering
    \includegraphics[width=1.0\textwidth]{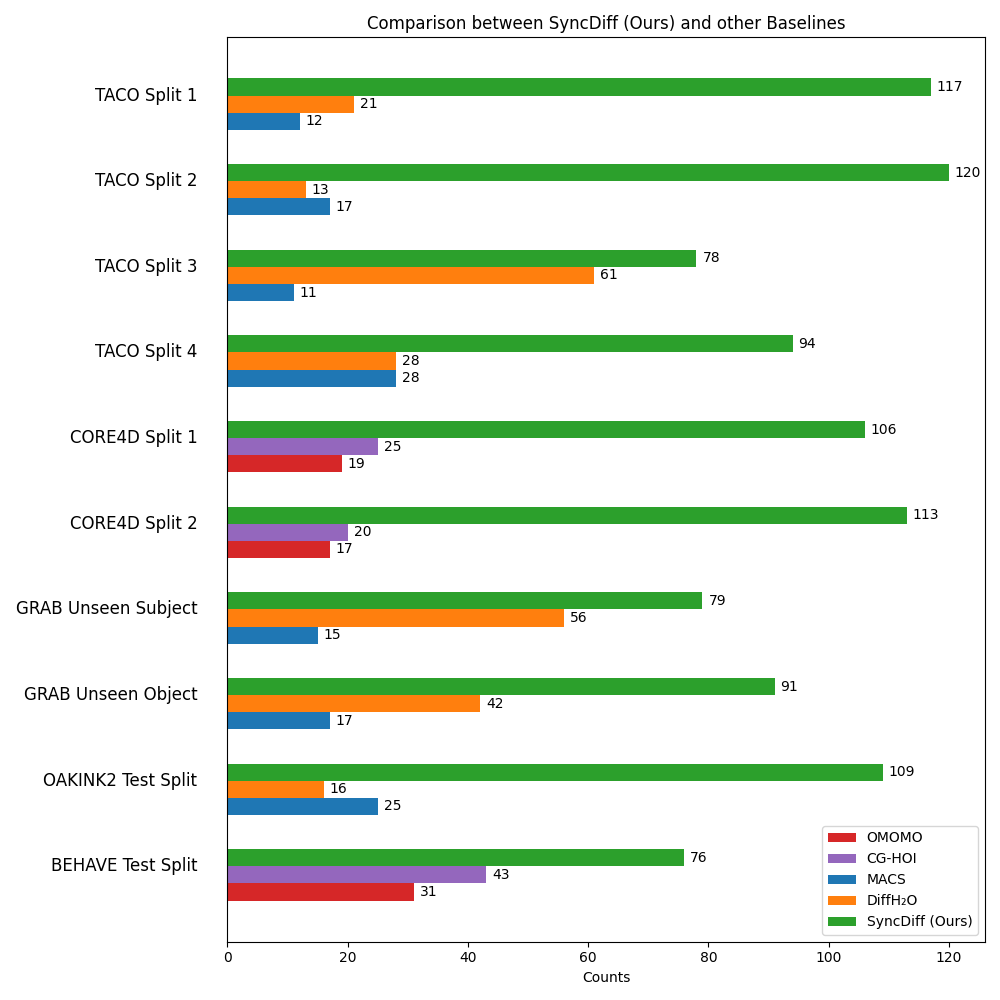}
    \caption{User study results on different dataset splits.}
    \label{fig:userstudy}
\end{figure*}

\begin{table}[t]
\centering
\small
\addtolength{\tabcolsep}{-3pt}
{
\begin{tabular}{c c c c c}
\toprule
\multirow{2}{*}{Method} & {CSIoU} & {IV} & {FID} & {RA} \\
& (\%, $\uparrow$) & ($\text{cm}^3$, $\downarrow$) & ($\downarrow$) & (\%, $\uparrow$) \\
\midrule
Ground-truth & 100.0 & 2.51  & 0.00 & 82.57\\

MACS~\cite{shimada2023macsmassconditioned3d} & 57.52 & 10.52  & 4.96 & 54.91\\

DiffH$_{2}$O~\cite{christen2024diffh2odiffusionbasedsynthesishandobject} & 55.50 & 5.59 & 5.18 & 50.48 \\

\textbf{SyncDiff (Ours)} & \textbf{72.14} & \textbf{4.41} & \textbf{2.65} & \textbf{74.83}\\

\midrule
w/o all & 62.96 & 6.73 & 6.63 & 48.96 \\
w/o decompose & 68.16 & 4.90 & 4.46 & 55.05 \\
w/o $\mathcal{L}_{\text{align}}$, exp sync & 57.59 & 8.94 & 3.82 & 70.54 \\
w/o $\mathcal{L}_{\text{align}}$ & 67.44 & 5.34 & 3.76 & 70.82 \\
w/o exp sync & 58.05 & 7.66 & 3.58 & 69.16 \\
\bottomrule
\end{tabular}
}
\vspace{-0.2cm}
\caption{Results on OAKINK2~\cite{zhan2024oakink2datasetbimanualhandsobject} dataset. The best in each column is highlighted in bold.}
\vspace{-0.8cm}
\label{tab:table3}
\end{table}

\textbf{Removal of $\mathcal{L}_{\text{align}}$ or Explicit Synchronization.} As is shown in Figures~\ref{fig:OAKINK2} and~\ref{fig:CORE4D}, removing either one of $\mathcal{L}_{\text{align}}$ or explicit synchronization leads to unreasonable penetration or contact loss between objects or between humans and objects, with explicit synchronization playing a more significant role. This phenomenon is also revealed in quantitative evaluation results from Tables~\ref{tab:table1},~\ref{tab:table2} and~\ref{tab:table3}. An observed phenomenon is that simply incorporating frequency decomposition (w/o $\mathcal{L}_{\text{align}}$, exp sync) poses higher demand on synchronization, which is manifested as worse contact-based metrics (Defined in Section~\ref{sec:4.2}) than ``w/o all". Simply integrating $\mathcal{L}_{\text{align}}$ (w/o exp sync) can not fully solve this issue, making explicit synchronization steps become a must.

\subsection{User Study}
\label{sec:4.5}
In order to evaluate semantic correctness of synthesized interactions from a perceptual aspect, we conduct a user study, comparing our method with all other baselines on the five datasets.

There are $10$ different splits in our experiments (TACO splits 1-4, CORE4D splits 1-2, OAKINK2 test split, GRAB splits of unseen subjects/objects, BEHAVE test split). We randomly draw 15 samples per split, resulting in a total of $150$ different questions ($15 \times 10$). Each question requires participants to choose the best option among \textbf{Ours} and all other baselines, based primarily on ``Completion, Realism, and Naturalness". If the decision is difficult to make, participants are also instructed to consider ``additional misalignments such as interpenetration, contact loss, jitter". $150$ different individuals are involved, and each is assigned $10$ different questions, where the option orders are randomly shuffled. Our mechanism guarantees that each question is answered by exactly $10$ different agents. Figure~\ref{fig:userstudy} shows the times each method voted as the ``best" in each split.

From the data of User Study (which is also corroborated by other metrics such as contact-based metrics and metrics regarding semantics), it can be observed that for datasets with fewer bodies, like GRAB or BEHAVE, the gap between our method and the baselines is not as significant as in datasets with more bodies. This is because, as the number of bodies increases, the requirements for synchronization become higher, and the interaction patterns between rigid objects become more complex, necessitating explicit modeling of high-frequency components through frequency decomposition, as well as those synchronization mechanisms. This further highlights the necessity of our approach in the general task of synchronized multi-body interaction synthesis.

\section{Conclusions and Discussions}

This paper presents SyncDiff, a unified framework for synchronized motion synthesis of multi-body HOI interaction by estimating both data sample scores and alignment scores, and jointly optimizing sample and alignment likelihoods in inference. We also introduce a frequency-domain decomposition to better capture high-frequency motions with semantics. Experiments on five datasets demonstrate that SyncDiff can be adapted to multiple scenarios with any number of humans, hands, and rigid objects. Comparative experiments also demonstrate that in each specific setting, our method achieves better contact accuracy and action semantics than a range of state-of-the-art baselines. Limitations, potential solutions, extensions, and some discussions of SyncDiff are provided in Section~\ref{sec:E}.

{
    \small
    \bibliographystyle{ieeenat_fullname}
    \bibliography{main}
}

\clearpage
\Large \textbf{Appendix}
\normalsize
\appendix

\section{Details of Our Two Synchronization Mechanisms}
\label{sec:A}

\subsection{Diffusion Model Basics}
\label{sec:A.1}
Diffusion models~\cite{sohldickstein2015deepunsupervisedlearningusing} and its variants~\cite{song2020generativemodelingestimatinggradients, ho2020denoisingdiffusionprobabilisticmodels}, especially latent diffusion, have been widely applied in different tasks, such as high-resolution image generation~\cite{rombach2022high} or video generation~\cite{ho2022video}.
They simulate the data distribution by introducing a series of variables $\{x_i\}_{i=1}^{T}$ with different noise levels. The forward noise process can be represented as $$q(x_t|x_{t-1})=\mathcal{N}(x_t; \sqrt{1-\beta_t}x_{t-1}, \beta_tI),$$
where $0<\beta_t<1$. We can derive that $$x_t=\sqrt{\bar{\alpha}_t}x_0+\sqrt{1-\bar{\alpha}_t}\epsilon,$$ where $\epsilon\sim \mathcal{N}(0, I)$, and $\bar{\alpha}_t=\prod_{i=1}^t (1-\beta_i)$.

For the inference process, from $T$ that is large enough so that $p(x_{T})\approx \mathcal{N}(0, I)$, do reverse sampling step by step, following
$$p_{\theta}(x_{t-1}|x_t)=\mathcal{N}(x_{t-1}; \mu_{\theta}(x_t, t), \sigma_t^2I),$$
where $\mu_{\theta}$ is the mean distribution center predicted by network, and $\sigma_t^2$ is pre-defined constant variance. Then
$$x_{t-1}=\mu_{\theta}(x_t, t)+\sigma_t z(z\sim \mathcal{N}(0, I)).$$
Finally we can get $x_0$, which is the denoised sample.

\subsection{Complete Proof of Our Two Synchronization Mechanisms}
\label{sec:A.2}
The goal in this section consists of two aspects:

1) Illustrate how to define a set of alignment scores featuring synchronization analogous to the commonly used data sample scores in diffusion models, and derive the corresponding loss term $\mathcal{L}_{\text{align}}$ in Section~\ref{sec:3.5};

2) Prove that the explicit synchronization formulas in Section~\ref{sec:3.5} are equivalent to maximum total likelihood sampling on the newly computed Gaussian distribution, where data sample scores and alignment scores are jointly considered.

Before we start, let's derive some commonly used formulas in diffusion models that we will need in our proof, which the readers might not be familiar with. If you are familiar with the step-by-step denoising formula of diffusion, you can skip directly to Eq ~\ref{eq:1} and ~\ref{eq:2}.

Suppose the noise-adding process has a total of $T$ steps, with each step's amplitude denoted by $\beta_t(t\in [T])$, we define $\alpha_t=1-\beta_t$, and $\bar{\alpha}_t=\prod_{t=1}^T \alpha_t$.

Then by basic principles of diffusion models, we have
$$x_t\sim \mathcal{N}(\sqrt{\alpha_t}x_{t-1}, (1-\alpha_t)I)$$
$$x_{t-1}\sim \mathcal{N}(\sqrt{\bar{\alpha}_{t-1}}x_0, (1-\bar{\alpha}_{t-1})I)$$
$$x_t\sim \mathcal{N}(\sqrt{\bar{\alpha}_t}x_0, (1-\bar{\alpha}_t)I)$$

According to Bayesian's Formula,
$$q(x_{t-1}|x_t, x_0)=\frac{q(x_t|x_{t-1})q(x_{t-1}|x_0)}{q(x_t|x_0)}$$

Taking negative logarithmic,

\begin{align*}
    &-\log q(x_{t-1}|x_t, x_0)\\
    &= \frac{(x_t-\sqrt{\alpha_t}x_{t-1})^2}{2(1-\alpha_t)}+\frac{(x_{t-1}-\sqrt{
    \bar{\alpha}_{t-1}}x_0)^2}{2(1-\bar{\alpha}_{t-1})}\\&-\frac{(x_t-\sqrt{
    \bar{\alpha}_t}x_0)^2}{2(1-\bar{\alpha}_t)}+\text{Const}\\
    &=\left[\frac{\alpha_t}{2(1-\alpha_t)}+\frac{1}{2(1-\bar{\alpha}_{t-1})}\right]x_{t-1}^2\\&-2\left[\frac{\sqrt{\alpha}_t}{2(1-\alpha_t)}x_t+\frac{\sqrt{\bar{\alpha}_{t-1}}}{2(1-\bar{\alpha}_{t-1})}x_0\right]x_{t-1}+C(x_t, x_0)\\
    &\overset{\triangle}{=} Ax_{t-1}^2+Bx_{t-1}+C\\
    &=A\left(x_{t-1}+\frac{B}{2A}\right)^2+C^{'}\\
\end{align*}
so
  $$  x_{t-1}\sim \mathcal{N}\left(-\frac{B}{2A}, \frac{1}{2A}I\right)\overset{\triangle}{=} \mathcal{N}(\mu_t, \sigma_t^2I)$$

where

\begin{align*}
    &\mu_t=-\frac{B}{2A}\\
    &=\frac{\frac{\sqrt{\alpha_t}}{{1-\alpha_t}}x_t+\frac{\sqrt{\bar{\alpha}_{t-1}}}{1-\bar{\alpha}_{t-1}}x_0}{\frac{\alpha_t}{1-\alpha_t}+\frac{1}{1-\bar{\alpha}_{t-1}}}\\
    &=\frac{\sqrt{\alpha}_t(1-\bar{\alpha}_{t-1})}{\alpha_t(1-\bar{\alpha}_{t-1})+1-\alpha_t}x_t+\frac{\sqrt{\bar{\alpha}_{t-1}}(1-\alpha_t)}{\alpha_t(1-\bar{\alpha}_{t-1})+(1-\alpha_t)}x_0\\
    &=\frac{\sqrt{\alpha}_t(1-\bar{\alpha}_{t-1})}{1-\bar{\alpha}_t}x_t+\frac{\sqrt{\bar{\alpha}_{t-1}}(1-\alpha_t)}{1-\bar{\alpha}_t}x_0
\end{align*}
Since $x_0=\frac{1}{\sqrt{\bar{\alpha}_t}}x_t-\frac{\sqrt{1-\bar{\alpha}_t}}{\sqrt{\bar{\alpha}_t}}\epsilon$,

\begin{equation}
\label{eq:1}
\begin{aligned}
    \mu_t &= \left[\frac{\sqrt{\alpha_t}(1-\bar{\alpha}_{t-1})}{1-\bar{\alpha}_t} + \frac{1-\alpha_t}{\sqrt{\alpha_t}(1-\bar{\alpha}_t)}\right] x_t 
    - \frac{1-\alpha_t}{\sqrt{\alpha_t(1-\bar{\alpha}_t)}} \epsilon \\
    &= \frac{1}{\sqrt{\alpha_t}}\left(x_t - \frac{1-\alpha_t}{\sqrt{1-\bar{\alpha}_t}}\epsilon\right)
\end{aligned}
\end{equation}

\begin{equation}
\label{eq:2}
\begin{aligned}
    \sigma_t^2 &= \frac{1}{2A} \\
    &= \frac{1}{\frac{\alpha_t}{1-\alpha_t} + \frac{1}{1-\bar{\alpha}_{t-1}}} \\
    &= \frac{1-\alpha_t-\bar{\alpha}_{t-1}-\bar{\alpha}_t}{1-\bar{\alpha}_t} \\
    &= \frac{(1-\alpha_t)(1-\bar{\alpha}_{t-1})}{1-\bar{\alpha}_t} \\
    &= \beta_t \frac{1-\bar{\alpha}_{t-1}}{1-\bar{\alpha}_t}
\end{aligned}
\end{equation}

To define alignment scores and derive $\mathcal{L}_{\text{align}}$ from it, we first need to review how traditional diffusion models derive the reconstruction loss term from data sample scores. In fact, due to the overly simple form of the reconstruction loss, its profound mathematical background is often overlooked. Like most generative models, the essence of the reconstruction loss lies in optimizing the \textbf{negative log-likelihood}$-\log p_{\theta}(x_0)$, where $\theta$ is the model parameters, and $p_{\theta}(x_0)$ is the probability of the model reconstructing the data $x_0$. However, due to the step-by-step denoising mechanism for diffusion models, it is challenging to directly optimize $-\log p_{\theta}(x_0)$. The common approach is to optimize the ELBO (Evidence Lower Bound)~\cite{sohldickstein2015deepunsupervisedlearningusing}. Note that

\begin{equation}
\begin{aligned}
    &-\mathbb{E}_{q(x_0)} \log p_{\theta}(x_0) \\
    &= -\mathbb{E}_{q(x_0)} \log \left( \int p_{\theta}(x_{0:T}) \text{d}x_{1:T} \right) \\
    &= -\mathbb{E}_{q(x_0)} \log \left( \int q(x_{1:T} | x_0) \frac{p_{\theta}(x_{0:T})}{q(x_{1:T} | x_0)} \text{d}x_{1:T} \right) \\
    &= -\mathbb{E}_{q(x_0)} \log \left( \mathbb{E}_{q(x_{1:T} | x_0)} \frac{p_{\theta}(x_{0:T})}{q(x_{1:T} | x_0)} \right) \\
    &\leq -\mathbb{E}_{q(x_{0:T})} \log \frac{p_{\theta}(x_{0:T})}{q(x_{1:T} | x_0)} \\
    &= \mathbb{E}_{q(x_{0:T})} \left[ \log \frac{q(x_{1:T} | x_0)}{p_{\theta}(x_{0:T})} \right]\\
    &=\mathbb{E}_{q(x_{0:T})}\left[\text{log}\frac{q(x_T|x_0)\prod_{t=2}^Tq(x_{t-1}|x_t, x_0)}{p_{\theta}(x_T)\prod_{t=1}^Tp_{\theta}(x_{t-1}|x_t)}\right]\\
    &=\mathbb{E}_{q(x_{0:T})}\left[\text{log}\frac{q(x_T|x_0)}{p_{\theta}(x_T)}\right. \\
    &\quad\left. +\sum\limits_{t=2}^T\text{log}\frac{q(x_{t-1}|x_t, x_0)}{p_{\theta}(x_{t-1}|x_t)}-\text{log }p_{\theta}(x_0|x_1)\right]\\
    &=\mathbb{E}_q\left[
    D_{\text{KL}}(q(x_T|x_0) \| p_{\theta}(x_T)) - \text{log }p_{\theta}(x_0|x_1) \right. \\
    &\quad\left. + \sum\limits_{t=2}^T D_{\text{KL}}(q(x_{t-1}|x_t, x_0) \| p_{\theta}(x_{t-1}|x_t)) \right]
\end{aligned}
\end{equation}

Our loss is primarily the error between the distribution $p_{\theta}$ predicted by the model during the reverse denoising process and the true distribution $q$. It is well known that the KL divergence between two Gaussian distributions $\mathcal{N}(\mu_1, \sigma_1^2)$, $\mathcal{N}(\mu_2, \sigma_2^2)$ is given by

\begin{equation}
\begin{aligned}
&D_{\text{KL}}(\mathcal{N}(\mu_1, \sigma_1^2)||\mathcal{N}(\mu_2, \sigma_2^2))\\
&\quad=\log\left(\frac{\sigma_2}{\sigma_1}\right)+\frac{\sigma_1^2+\Vert\mu_1-\mu_2\Vert_2^2}{2\sigma_2^2}-\frac{1}{2}
\end{aligned}
\end{equation}

In our derivation, $q(x_{t-1}\mid x_t, x_0)$ means the ground-truth reverse process distribution, while $p_{\theta}(x_{t-1}\mid x_t)$ is our predicted distribution in stepwise denoising. Their mean values are referred to as $\mu_t$ and $\hat{\mu}_t$, and the standard variance $\sigma_t$ is a predefined constant in DDPM~\cite{ho2020denoisingdiffusionprobabilisticmodels}. Plug in $q(x_{t-1}\mid x_t, x_0)=\mathcal{N}(\mu_t, \sigma_t^2)$, $p_{\theta}(x_{t-1}\mid x_t)=\mathcal{N}(\hat{\mu}_t, \sigma_t^2)$, we have

$$D_{\text{KL}}(q(x_{t-1}|x_t, x_0) \| p_{\theta}(x_{t-1}|x_t))=\frac{1}{2\sigma_t^2}\Vert \hat{\mu}_t-\mu_t\Vert_2^2$$

In general, diffusion models predict the noise $\epsilon$, but in our setting, we need to define the alignment loss later. Thus, it's more convenient to direct predict $\hat{x}_0$, which is the result after complete denoising. Note that

$$\epsilon=\frac{1}{\sqrt{1-\bar{\alpha}_t}}(x_t-\sqrt{\bar{\alpha}_t}x_0),$$

so

\begin{equation}
\begin{aligned}
    \mu_t &=\frac{1}{\sqrt{\alpha}_t}\left(x_t-\frac{1-\alpha_t}{\sqrt{1-\bar{\alpha}_t}}\epsilon\right)\\
    &=\frac{1}{\sqrt{\alpha}_t}\left(x_t-\frac{(1-\alpha_t)(x_t-\sqrt{\bar{\alpha}_t}x_0)}{1-\bar{\alpha}_t}\right)\\
    &=\frac{1}{\sqrt{\alpha}_t}\left(\frac{\alpha_t(1-\bar{\alpha}_{t-1})}{1-\bar{\alpha}_t}x_t+\frac{(1-\alpha_t)\sqrt{\bar{\alpha}_t}}{1-\bar{\alpha}_t}x_0\right)\\
    &=\frac{\sqrt{\alpha_t}(1-\bar{\alpha}_{t-1})}{1-\bar{\alpha}_t}x_t+\frac{\sqrt{\bar{\alpha}_{t-1}}\beta_t}{1-\bar{\alpha}_t}x_0.
\end{aligned}
\end{equation}

Similarly,
$$\hat{\mu}_t=\frac{\sqrt{\alpha_t}(1-\bar{\alpha}_{t-1})}{1-\bar{\alpha}_t}x_t+\frac{\sqrt{\bar{\alpha}_{t-1}}\beta_t}{1-\bar{\alpha}_t}\hat{x}_0.$$

Therefore,

$$D_{\text{KL}}(q(x_{t-1}|x_t, x_0) \| p_{\theta}(x_{t-1}|x_t))=\frac{\bar{\alpha}_{t-1}\beta_t^2}{2\sigma_t^2(1-\bar{\alpha}_t)^2}\Vert x_0-\hat{x}_0\Vert_2^2.$$

By removing the coefficient, it becomes our simple reconstruction loss. Note that in our implementation, we split it into $\mathcal{L}_{\text{dc}}$ and $\mathcal{L}_{\text{ac}}$ for separate supervision.
Similarly, in order to derive the alignment loss, our approach is still to express the loss function as some form of negative log-likelihood. Consider a triplet $(\hat{x}_{ \textbf{b}_1}, \hat{x}_{ \textbf{b}_2}, \hat{x}_{ \textbf{b}_2\rightarrow \textbf{b}_1})$, where $\textbf{b}_1$, $\textbf{b}_2$ are two different bodies. We know that the definition of the score function is $\nabla \log p(x)$, which is just the negative gradient of the negative log-likelihood. As stated in the introduction part, we hope that this score function can guide $\hat{x}_{ \textbf{b}_2\rightarrow \textbf{b}_1}$ towards $\text{rel}(\hat{x}_{ \textbf{b}_1}, \hat{x}_{ \textbf{b}_2})$. A simple solution is to let $\hat{x}_{ \textbf{b}_2\rightarrow \textbf{b}_1}$ follow a distribution $\mathcal{N}(\mu, \sigma^2)$, where $\mu=\text{rel}(\hat{x}_{ \textbf{b}_1}, \hat{x}_{ \textbf{b}_2})$, and $\sigma$ is a parameter that we can tune. Note that a vector $\hat{x}$ satisfying $\hat{x}\sim \mathcal{N}(\hat{\mu}, \sigma^2I)$ has the probability density function

$$p(\hat{x})=\frac{1}{(2\pi)^{d/2}|\sigma^2|^{1/2}}\exp\left(-\frac{1}{2}(\hat{x}-\hat{\mu})^\top \left(\sigma^2\right)^{-1}(\hat{x}-\hat{\mu})\right)$$
where $d$ is the dimention of $\hat{x}$. The negative log-likelihood of $p(\hat{x})$, $-\log p(\hat{x})$, can be written as

$$-\log p(\hat{x})=\frac{1}{2\sigma^2}\Vert \hat{x}-\hat{\mu}\Vert_2^2$$

It is not difficult to obtain that the negative log-likelihood is $\frac{1}{2\sigma^2}\Vert \hat{x}_{ \textbf{b}_2\rightarrow \textbf{b}_1}-\text{rel}(\hat{x}_{ \textbf{b}_1}, \hat{x}_{ \textbf{b}_2})\Vert_2^2$. Summing up for all such pairs $(\textbf{b}_1, \textbf{b}_2)$, and removing the coefficients, we can derive the alignment loss

\begin{equation}
\begin{aligned}
\mathcal{L}_{\text{align}}= & \sum\limits_{j_1, j_2\in [1, m], j_1\neq j_2}\Vert \hat{x}_{ o_{j_2}\rightarrow o_{j_1}}-\text{rel}(\hat{x}_{ o_{j_1}}, \hat{x}_{ o_{j_2}})\Vert_2^2 \\
& +\sum\limits_{i\in [1,n], j\in [1,m]}\Vert \hat{x}_{ h_i\rightarrow o_j}-\text{rel}(\hat{x}_{ o_j}, \hat{x}_{ h_i})\Vert_2^2.
\end{aligned}
\end{equation}

Next, let's prove the equivalence between the explicit synchronization formulas and maximum total likelihood sampling in inference. In our task, during the inference process, suppose we want to derive $\hat{x}^{'}=\hat{x}_{t-1}$ from $\hat{x}=\hat{x}_t$, we first use the denoising backbone to predict the mean value of distribution $\hat{\mu}=\hat{\mu}_t$. 

In our task, $\hat{x}^{'}=[\hat{x}^{'}_{ o_1}, \hat{x}^{'}_{ o_2}, \dots, \hat{x}^{'}_{ o_m},  \hat{x}^{'}_{ h_1}, \hat{x}^{'}_{ h_2}, \dots, \hat{x}^{'}_{ h_n}, \\ \hat{x}^{'}_{ o_1\rightarrow o_2}, \hat{x}^{'}_{ o_1\rightarrow o_3}, \dots, \hat{x}^{'}_{ o_m\rightarrow o_{m-1}},  \hat{x}^{'}_{ h_1\rightarrow o_1}, \hat{x}^{'}_{ h_1\rightarrow o_2}, \dots, \\ \hat{x}^{'}_{ h_n\rightarrow o_m}]$, and $\hat{\mu}=[\hat{\mu}_{ o_1}, \hat{\mu}_{ o_2}, \dots, \hat{\mu}_{ o_m},  \hat{\mu}_{ h_1}, \hat{\mu}_{ h_2}, \dots, \hat{\mu}_{ h_n}, \\ \hat{\mu}_{ o_1\rightarrow o_2}, \hat{\mu}_{ o_1\rightarrow o_3}, \dots, \hat{\mu}_{ o_m\rightarrow o_{m-1}},  \hat{\mu}_{ h_1\rightarrow o_1}, \hat{\mu}_{ h_1\rightarrow o_2}, \dots, \\ \hat{\mu}_{ h_n\rightarrow o_m}]$, so

\begin{align*}
    \frac{1}{2\sigma^2}\Vert \hat{x}^{'}-\hat{\mu}\Vert_2^2 &= \frac{1}{2\sigma^2}\left(\sum\limits_{i=1}^n \Vert \hat{x}^{'}_{ h_i}-\hat{\mu}_{ h_i}\Vert_2^2 \right.\\ 
    &\quad + \sum\limits_{j=1}^m \Vert \hat{x}^{'}_{ o_j}-\hat{\mu}_{ o_j}\Vert_2^2 \\
    &\quad + \sum\limits_{j_1, j_2\in [1, m], j_1\neq j_2} \Vert \hat{x}^{'}_{ o_{j_2}\rightarrow o_{j_1}}-\hat{\mu}_{ o_{j_2}\rightarrow o_{j_1}}\Vert_2^2 \\
    &\quad + \left. \sum\limits_{i\in [1, n], j\in [1, m]} \Vert \hat{x}^{'}_{ h_i\rightarrow o_j}-\hat{\mu}_{ h_i\rightarrow o_j}\Vert_2^2\right)
\end{align*}

On the other side, the negative logarithm of alignment likelihood is defined as
\begin{align*}
    \mathcal{P}_{\text{align}}&=\sum\limits_{v=1}^{|V|}\lambda_v \Vert c_v- a_v \circ b_v\Vert _2^2
\end{align*}
where $\lambda_{v\in [1, V]}$ are a hyperparameters, $\{(a_v, b_v, c_v)\}_{v=1}^{|V|}$ is the set consisting of all triplets $(\hat{x}^{'}_{ \textbf{b}_1}, \hat{x}^{'}_{ \textbf{b}_2}, \hat{x}^{'}_{ \textbf{b}_2\rightarrow \textbf{b}_1})$, where $c_v$ can be computed by $a_v$ and $b_v$ through combination operation or relative operation. For example, if $a_v=\hat{x}^{'}_{ o_1}$ and $b_v=\hat{x}^{'}_{ o_2}$, then $c_v=\hat{x}^{'}_{ o_2\rightarrow o_1}$, and $a_v\circ b_v=\text{rel}\left(\hat{x}^{'}_{ o_1}, \hat{x}^{'}_{ o_2}\right)$. If $a_v=\hat{x}^{'}_{ o_1}$ and $b_v=\hat{x}^{'}_{ h_1\rightarrow o_1}$, then $c_v=\hat{x}^{'}_{ h_1}$, and $a_v\circ b_v=\text{comb}\left(\hat{x}^{'}_{ o_1}, \hat{x}^{'}_{ h_1\rightarrow o_1}\right)$. The alignment likelihoods encompass the likelihoods of all binary computational relationships. Note that the alignment negative log-likelihood here is different from alignment loss $\mathcal{L}_{\text{align}}$ in the main text, which only considers relative operations rel, without combination operations comb. The intrinsic mathematical meaning of one item $\lambda_v\Vert c_v-a_v\circ b_v\Vert ^2$ is similar to the above derivations of $\mathcal{L}_{\text{align}}$, where we let $c_v$ follow a Gaussian distribution with a mean of $\hat{\mu}=a_v\circ b_v$ and a variance of $\sigma^2=\frac{1}{2\lambda_v}$. 

Now consider fixing some $c_v=\hat{x}^{''}$ (This is the motion of one single body, which is different from $\hat{x}$ and $\hat{x}^{'}$, the representation comprising of all individual/relative motions). It might be computed by $K$ pairs of $(a_k, b_k)$. Take $\hat{x}^{''}=\hat{x}^{'}_{ o_1}$ as an example. Here $K=m-1$, and $(a_1, b_1)=(\hat{x}^{'}_{ o_2}, \hat{x}^{'}_{ o_1\rightarrow o_2})$, $(a_2, b_2)=(\hat{x}^{'}_{ o_3}, \hat{x}^{'}_{ o_1\rightarrow o_3})$, $\dots$, $(a_{m-1}, b_{m-1})=(\hat{x}^{'}_{ o_m}, \hat{x}^{'}_{ o_1\rightarrow o_m})$. We need to simultaneously make $\hat{x}^{''}$ as close as possible to the corresponding part $\hat{\mu}^{''}=\hat{\mu}_{ o_1}$ in $\hat{\mu}$ predicted by the diffusion model, while ensuring that $\hat{x}^{''}$ aligns with each predicted pair $(a_k,b_k)$. Add these terms together, maximizing total likelihood (the combination of data sample likelihood and alignment likelihoods) is equivalent to minimizing

$$\mathcal{P}_{\hat{x}^{''}}=\frac{1}{2\sigma^2}\Vert \hat{x}^{''}-\hat{\mu}^{''}\Vert_2^2+\sum\limits_{k=1}^{K} \lambda_k \Vert \hat{x}^{''}- a_k \circ b_k\Vert _2^2$$

A problem here is that $\hat{\mu}$ is predicted by the model based on the result of step $t$, but $(a_1,b_1)$, $\dots$, $(a_{K},b_{K})$ all belong to step $(t-1)$ along with $\hat{x}^{''}$. Here, we make an assumption that $\hat{x}_t$ and $\hat{x}_{t-1}$ are close, so that we can take $(a_1,b_1)$, $(a_2, b_2)$, $\dots$, $(a_{K}, b_{K})$ from $\hat{x}_t=\hat{x}$.

Denote $\hat{\mu}^{''}$ as $f_0$, and $a_1\circ b_1$, $a_2\circ b_2$, $\dots$, $a_{K}\circ b_{K}$ as $f_1$, $f_2$, $\dots$, $f_{K}$. Here $f_0$, $f_1$, $\dots$, $f_{K}$ are all deterministic values calculated from some certain parts of $\hat{x}_t$. Also let $\lambda_0=\frac{1}{2\sigma^2}$, then
\begin{align*}
    \mathcal{P}_{\hat{x}^{\prime\prime}}&=\sum\limits_{k=0}^{K} \lambda_k \Vert \hat{x}^{\prime\prime}-f_k\Vert _2^2\\
    &=\sum\limits_{k=0}^{K} \lambda_k \left(\hat{x}^{\prime\prime \top} \hat{x}^{\prime\prime}-2f_k^\top \hat{x}^{\prime\prime}+f_k^\top f_k\right)\\
    &=\left(\sum\limits_{k=0}^{K} \lambda_k\right) \Vert \hat{x}^{\prime\prime}\Vert _2^2-2\left(\sum\limits_{k=0}^{K} \lambda_k f_k\right)^\top \hat{x}^{\prime\prime}+\sum\limits_{k=0}^{K} \lambda_k \Vert f_k\Vert _2^2
\end{align*}

This can be viewed as the negative log-likelihood of a new Gaussian distribution 
\(\hat{x}^{\prime\prime} \sim \mathcal{N}(\hat{\mu}^{\prime}, \sigma^{\prime 2})\), where

\[
\hat{\mu}^{\prime} = \sum\limits_{k=0}^{K} \frac{\lambda_k}{\sum\limits_{k=0}^{K} \lambda_k} f_k
\]

\[
\sigma^{\prime 2} = \frac{1}{2 \left(\sum\limits_{k=0}^{K} \lambda_k \right)}
\]

Finally, it's time to consider the specific body types for calculation. 

1. \textbf{For individual motions of rigid body} $o_j(j\in [1, m ])$ (Here we assume that $m>1$, otherwise there is no need for explicit synchronization on this part), relevant pairs of $(a_k, b_k)$ consist of $(\hat{x}_{ o_{j^{'}}}, \hat{x}_{ o_j\rightarrow o_{j^{'}}})(j^{'}\neq j)$. $\lambda_0=\frac{1}{2\sigma^2}$, $\lambda_1=\lambda_2=\dots=\lambda_{m-1}=\frac{\overline{\lambda}}{m-1}$. Here $\overline{\lambda}$ is an empirical value, satisfying
$$\overline{\lambda}=\frac{\lambda_{\text{exp}}}{R}\sum\limits_{r=1}^{R} \frac{1}{2\sigma_{t_r}^2}$$
where $1\leq t_1<t_2<\dots<t_{R}\leq T$ are the synchronization timesteps, and $\sigma_{t_1}$, $\sigma_{t_2}$, $\dots$, $\sigma_{t_{R}}$ are the original correspondent standard variances(without synchronization). The value of hyperparameter $\lambda_{\text{exp}}$ can be found in Table ~\ref{tab:hyperparameter}. Therefore,
\begin{align*}
    &\hat{\mu}^{'}_{ o_j}\\
    &=\frac{1}{\frac{1}{2\sigma^2}+\overline{\lambda}}\left(\lambda_0 \hat{\mu}_{ o_j}+\sum\limits_{j^{'}\neq j}\frac{\overline{\lambda}}{m-1}\text{comb}\left(\hat{x}_{ o_{j^{'}}}, \hat{x}_{ o_j\rightarrow o_{j^{'}}}\right)\right)\\
    &=\frac{1}{1+2\sigma^2\overline{\lambda}}\hat{\mu}_{ o_j}+\frac{\frac{2}{m-1}\sigma^2\overline{\lambda}}{1+2\sigma^2\overline{\lambda}}\sum\limits_{j^{'}\neq j}\text{comb}\left(\hat{x}_{ o_{j^{'}}}, \hat{x}_{ o_j\rightarrow o_{j^{'}}}\right)
\end{align*}

2. \textbf{For individual motions of articulated} skeleton $h_i(i\in [1, n])$, relevant pairs of $(a_k, b_k)$ consist of $(\hat{x}_{ o_j}, \hat{x}_{ h_i\rightarrow o_j})(j\in [1, m])$. $\lambda_0=\frac{1}{2\sigma^2}$, $\lambda_1=\lambda_2=\dots=\lambda_m=\frac{\overline{\lambda}}{m}$. Therefore,
\begin{align*}
    \hat{\mu}^{'}_{ h_i}&=\frac{1}{\frac{1}{2\sigma^2}+\overline{\lambda}}\left(\lambda_0 \hat{\mu}_{ h_i}+\sum\limits_{j\in [1, m]}\frac{\overline{\lambda}}{m}\text{comb}\left(\hat{x}_{ o_j}, \hat{x}_{ h_i\rightarrow o_j}\right)\right)\\
    &=\frac{1}{1+2\sigma^2\overline{\lambda}}\hat{\mu}_{ o_j}+\frac{\frac{2}{m}\sigma^2\overline{\lambda}}{1+2\sigma^2\overline{\lambda}}\sum\limits_{j\in [1, m]}\text{comb}\left(\hat{x}_{ o_j}, \hat{x}_{ h_i\rightarrow o_j}\right)
\end{align*}

3. \textbf{For relative motions}, there is only one relevant pair of $(a_k, b_k)$, where $a_k$ and $b_k$ are both individual motions, which can obtain the relative motion through relative composition. Here $\lambda_0=\frac{1}{2\sigma^2}$, $\lambda_1=\overline{\lambda}$. Therefore,
\begin{align*}
    \hat{\mu}^{'}_{ o_j\rightarrow o_{j^{'}}}&=\frac{1}{\frac{1}{2\sigma^2}+\overline{\lambda}}\left(\lambda_0 \hat{\mu}_{ o_j\rightarrow o_{j^{'}}}+\lambda_1\text{rel}\left(\hat{x}_{ o_{j^{'}}}, \hat{x}_{ o_j}\right)\right)\\
    &=\frac{1}{1+2\sigma^2\overline{\lambda}}\hat{\mu}_{ o_j\rightarrow o_{j^{'}}}+\frac{2\sigma^2\overline{\lambda}}{1+2\sigma^2\overline{\lambda}}\text{rel}\left(\hat{x}_{ o_{j^{'}}}, \hat{x}_{ o_j}\right)
\end{align*}

\begin{align*}
    \hat{\mu}^{'}_{ h_i\rightarrow o_j}&=\frac{1}{\frac{1}{2\sigma^2}+\overline{\lambda}}\left(\lambda_0 \hat{\mu}_{ h_i\rightarrow o_j}+\lambda_1\text{rel}\left(\hat{x}_{ o_j}, \hat{x}_{ h_i}\right)\right)\\
    &=\frac{1}{1+2\sigma^2\overline{\lambda}}\hat{\mu}_{ h_i\rightarrow o_j}+\frac{2\sigma^2\overline{\lambda}}{1+2\sigma^2\overline{\lambda}}\text{rel}\left(\hat{x}_{ o_j}, \hat{x}_{ h_i}\right)
\end{align*}

For the derivation of $\hat{x}^{'}$, we only need to add noise $\sigma^{'}\epsilon(\epsilon\sim \mathcal{N}(0, I))$ to $\hat{\mu}^{'}$, where

$$\sigma^{'}=\sqrt{\frac{1}{2\left(\frac{1}{2\sigma^2}+\overline{\lambda}\right)}}=\sqrt{\frac{\sigma^2}{1+2\sigma^2\overline{\lambda}}}$$

Thus, we have completed the proof. $\square$
\subsection{Algorithm for Explicit Synchronization}
\label{sec:A.3}
To help readers better understand the process of explicit synchronization in inference, we have specially prepared Algorithm~\ref{alg:1} here.

\begin{algorithm}
\caption{Explicit Synchronization}
\label{alg:1}
\begin{algorithmic}[1]
\Procedure{$\text{Exp}\_\text{Sync}$}{cond}
    \State $\lambda \gets 0$, $R \gets 0$
    \For{$t \gets T$ \textbf{to} $1$}
        \If{$t \bmod s = \lfloor s / 2 \rfloor$}
            \State $\lambda \gets \lambda + \frac{1}{2\sigma_t^2}$, $R \gets R + 1$
        \EndIf
    \EndFor
    \State $\overline{\lambda} \gets \lambda_{\text{exp}}\cdot\frac{\lambda}{R}$
    \State $\hat{x}_T \sim \mathcal{N}(0, I)$
    
    \For{$t \gets T$ \textbf{to} $1$}
        \State $\hat{\mu} \gets \text{Denoise}(\hat{x}_t)$
        \If{$t \bmod s = \lfloor s / 2 \rfloor$}
            \State $\lambda_0 \gets \frac{1}{1 + 2\sigma_t^2\overline{\lambda}}$, $\lambda_1 \gets \frac{2\sigma_t^2\overline{\lambda}}{1 + 2\sigma_t^2\overline{\lambda}}$
            \State $\hat{x}^{'} \gets \lambda_0 \cdot \hat{\mu}$
            \For{$j \gets 1$ \textbf{to} $m$}
                \If{$m = 1$}
                    \State $\hat{x}^{'}_{ o_j} \gets \hat{x}^{'}_{ o_j} + \lambda_1 \cdot \hat{\mu}_{ o_j}$
                \Else
                    \For{$j^{'} \neq j$}
                        \State $\hat{x}^{'}_{ o_j} \gets \hat{x}^{'}_{ o_j} + \frac{\lambda_1}{m-1} \cdot \text{comb}\left(\hat{x}_{ o_{j^{'}}}, \hat{x}_{ o_j\rightarrow o_{j^{'}}}\right)$
                        \State $\hat{x}^{'}_{ o_j\rightarrow o_{j^{'}}} \gets \hat{x}^{'}_{ o_j\rightarrow o_{j^{'}}} + \lambda_1 \cdot \text{rel}\left(\hat{x}_{ o_{j^{'}}},\hat{x}_{ o_j}\right)$
                    \EndFor
                    
                \EndIf
            \EndFor
            \For{$i \gets 1$ \textbf{to} $n$}
                \For{$j \gets 1$ \textbf{to} $m$}
                    \State $\hat{x}^{'}_{ h_i} \gets \hat{x}^{'}_{ h_i} + \frac{\lambda_1}{m} \cdot \text{comb}\left(\hat{x}_{ o_j}, \hat{x}_{ h_i\rightarrow o_j}\right)$
                    \State $\hat{x}^{'}_{ h_i\rightarrow o_j} \gets \hat{x}^{'}_{ h_i\rightarrow o_j} + \lambda_1 \cdot \text{rel}\left(\hat{x}_{ o_j}, \hat{x}_{ h_i}\right)$
                \EndFor
            \EndFor
            \State $\sigma^{'} \gets \sqrt{\frac{\sigma_t^2}{1 + 2\sigma_t^2\overline{\lambda}}}$
        \Else
            \State $\hat{x}^{'} \gets \hat{\mu}$
            \State $\sigma^{'} \gets \sigma_t$
        \EndIf
        \State $\epsilon \sim \mathcal{N}(0, I)$
        \State $\hat{x}^{'} \gets \hat{x}^{'} + \sigma^{'} \cdot \epsilon$
        \State $\hat{x}_{t-1}\gets \hat{x}^{'}$
    \EndFor
\EndProcedure
\end{algorithmic}
\end{algorithm}

\section{Supplementary Experiments}
\label{sec:B}

\subsection{Determine the cutoff boundary $L$ for Frequency Decomposition}
\label{sec:B.1}
In frequency decomposition of Section~\ref{sec:3.3}, we discard signals with frequencies higher than $\phi_L/2\pi$, where $\phi_L=L/N$, and $L$ is the cutoff boundary. We analyzed the impact of different $L$ on the maximum error $\epsilon$, which is averaged on the dimension of $N$ (number of frames) and takes the maximum across all degrees of freedom (the dimension of $D_{\text{sum}}$). The errors listed in Table ~\ref{tab:L} are averaged over all data samples in the datasets and are measured in \textbf{millimeters}.

\begin{table}[h!]
    \centering
    \footnotesize
    \begin{tabular}{c|c|c|c|c|c}
        \toprule
$L$ & TACO & CORE4D & OAKINK2 & GRAB & CG-HOI\\ \midrule
6  & 8.3 & 33.1 & 6.1 & 8.4 & 11.5\\ 
8  & 6.8 & 15.3   & 4.9 & 6.9 & 7.8\\ 
12 & 4.9 & 13.4 & 3.7   & 5.1 & 6.8\\ 
16 & 3.9 & 11.4 & 3.0    & 4.0 & 5.9\\ 
20 & 3.2  & 9.0 & 2.6  & 3.2  & 5.2\\ 
25 & 2.6  & 7.2 & 2.2  & 2.6  & 4.8\\ \bottomrule
    \end{tabular}
    \caption{Results for the impact of different $L$ on maximum error.}
    \label{tab:L}
\end{table}

Our goal is to minimize $L$, provided that the error between filtered and raw motions remains negligible relative to the original signal amplitude. To strike a balance between motion representation fidelity and simplicity, as well as filtering of too high-frequency noise in mocap datasets, in practice, we take $L=16$.

\subsection{Determine the Interval $s$ for Explicit Synchronization and Computational Cost for the Process}
\label{sec:B.2}
In explicit synchronization from Section~\ref{sec:3.6}, to balance inference speed and performance, we choose a hyperparameter $s(s<<T)$, where $T=1000$ is the total number of denoising steps, which means we only perform explicit synchronization operations every $s$ step, $R=T/s$ times in total.

We conduct experiments on TACO Split 1 using different $s$, as shown in Table ~\ref{tab:t0}. RA refers to recognition accuracy. To reduce computational cost without affecting the semantic accuracy of synthesized motions too much, in practice, we choose $s=50$.

\begin{table}[h!]
    \centering
    \begin{tabular}{c|c c}
        \toprule
        $s$               & Inference time per sample(s) & RA(\%, $\uparrow$)                                \\ 
        \midrule
        $1$ & $88.5$ & $74.04$                                     \\
        
        $5$ & $22.2$ & $73.91$                                     \\ 
        
        $10$ & $14.2$ & $74.09$                                     \\ 
        
        $50$ & $7.6$ & $73.28$                                     \\ 
        
        $100$ & $6.7$ & $72.15$ \\
        
        $500$ & $6.1$ & $70.52$ \\
        
        w/o exp & $5.9$ & $67.27$ \\
        \bottomrule
    \end{tabular}
    \caption{Results for different $s$ on TACO Split 1.}
    \label{tab:t0}
\end{table}

Here we also list the computational cost comparison between SyncDiff and each baselines. We test inference speed on a single NVIDIA A40 GPU for synthesizing a 200-frame sequence. Time consumption across different methods are shown in Tables ~\ref{tab:inference_time_human} and ~\ref{tab:inference_time_hand}.

\begin{table}[h!]
\centering
\vspace{-0.2cm}

\addtolength{\tabcolsep}{-3pt}
{
\begin{tabular}{|c|c|c|}
\hline  
\backslashbox{Method}{Dataset} & CORE4D & BEHAVE \\
\hline
OMOMO & 7.1s & 5.9s \\
CG-HOI & 10.7s & 7.6s \\
\textbf{SyncDiff (Ours)} & 6.5s & 4.2s \\
w/o exp sync & 5.4s & 3.6s \\
\hline

\end{tabular}
}
\caption{Inference time for a 200-frame human-object-interaction sequence.}
\label{tab:inference_time_human}
\end{table}

\begin{table}[h!]
\centering

\addtolength{\tabcolsep}{0pt}
{
\begin{tabular}{|c|c|c|c|}
\hline  
\backslashbox{Method}{Dataset} & TACO & OAKINK2 & GRAB \\
\hline
MACS & 7.4s & 7.0s & 5.7s \\
DiffH$_2$O & 8.6s & 8.9s & 7.4s \\
\textbf{SyncDiff (Ours)} & 7.7s & 7.3s & 6.6s \\
w/o exp sync & 5.9s & 5.6s & 5.4s \\
\hline
\end{tabular}
}
\vspace{0cm}
\caption{inference time for a 200-frame hand-object-interaction sequence.}
\label{tab:inference_time_hand}
\end{table}

It can be observed that computational cost of \textbf{SyncDiff} is no worse than most baselines, since specific designs like contact guidance (CG-HOI) or multi-stage synthesis (OMOMO / MACS / DiffH$_2$O) are circumvented. Comparison to ``w/o exp sync" shows that the exp sync operation applied every $s=50$ steps adds modest time cost of less than $25\%$.

\subsection{Check the Reconstruction Quality of High-frequency Components}
\label{sec:B.3}
From the comparison between \textbf{Ours} and ``w/o decompose" in the main text and supplementary videos, we can clearly see that if high-frequency components are not explicitly modeled, the motion trajectory tends to miss subtle high-frequency details, which are crucial for the accuracy of motion semantics. For example, the spatula merely contacts and gets stuck on the plate instead of moving back and forth to complete the \emph{scrape off} motion; similarly, when a person walks, his/her legs fail to alternate properly and instead slide forward like ``ice skating".

However, frequency decomposition consists of 1) the filter of too high-frequency noise, and 2) the explicit frequency domain representation ($x_{\text{F}}$) of high-frequency components with semantics ($x_{\text{ac}}$), where the former is relatively trivial. We need to decouple the contributions of them in further ablation studies. Additionally, a noteworthy phenomenon is that, compared to the baselines, our method exhibits less jitter (See videos of CORE4D/OAKINK2/GRAB ``comparison to baselines"), which could be attributed to either the filter of too high-frequency noise or the two synchronization mechanisms. It is essential to investigate the core source of this effect.

Identifying the source of less jitter is relatively straightforward. It is noted that the motion trajectories for training in the ``w/o decompose" scenario includes those too high-frequency noise (See Section~\ref{sec:C.4}). However, as observed in the ``w/o decompose" videos of TACO, there is almost no large-scale jitter in the synthesized motions, with the only issue being the absence of high-frequency interactions between objects. This is because, in the ``w/o decompose" experiments, both synchronization mechanisms are effective, ensuring that the generated motions are sufficiently aligned and thus reducing jitter. Conversely, in the OAKINK2 ablation studies of synchronization, whether it's ``w/o align loss" or ``w/o exp sync," the use of frequency decomposition also removes too high-frequency noise, yet it still results in greater jitter compared to the complete SyncDiff (Ours). This sufficiently demonstrates that the contribution to less jitter is primarily due to the synchronization mechanisms, rather than the filter of too high-frequency noise. A more straightforward method of verification is to observe that the errors in Table ~\ref{tab:L} caused by the filter of too high-frequency noise are almost negligible compared to the scales of the motions, indicating that this operation theoretically has minimal impact on the model's performance.

To decouple the effects of the frequency-domain based representation and the filter of too high-frequency noise from a more strict aspect, we conduct the following experiments. Besides ``w/o decompose", we add ``\textbf{simply filter}", which replaces $x_t$ in ``w/o decompose" with $x_{t, \text{dc}}+x_{t, \text{ac}}$, simulating the effect of filtering too high-frequency noise. ``\textbf{Only dc}" replaces $x_t$ with $x_{t, \text{dc}}$, simulating the effect of using no high-frequency signals (including those with semantics).

\textbf{Experiment 1:} On the four test splits of TACO, we only add noise for no larger than $200$ steps (where the full train pipeline needs $1000$ steps), and examine the reconstruction ability of different methods. For any predicted motion trajectory $\hat{x}$, we decompose it as described in Section~\ref{sec:3.3}, to obtain the high-frequency component $\hat{x}_{\text{ac}}$. We then examine the discrepancy between $\hat{x}_{\text{ac}}$ and the ground truth $x_{\text{ac}}$. The experimental results on the four splits of TACO are shown in Table ~\ref{tab:recon}, proving that explicitly representing high-frequency signals with semantics in the frequency domain indeed improves the effect of reconstruction.

\begin{table}[t]
\centering
\small
\addtolength{\tabcolsep}{-3pt}
{
\begin{tabular}{c|c c c c}
\toprule  
\multirow{2}{*}{Method} &  \multicolumn{4}{c}{Error (cm, $\downarrow$)} \\
\cline{2-5}
 & Test1 & Test2 & Test3 & Test4 \\
\midrule
MACS~\cite{shimada2023macsmassconditioned3d} & 2.43 & 2.37 & 3.11 & 2.74 \\

DiffH$_{2}$O~\cite{christen2024diffh2odiffusionbasedsynthesishandobject} & \textbf{1.74} & 2.08 & 3.12 & 3.31 \\

\textbf{SyncDiff (Ours)} & 1.99 & \textbf{1.63} & \textbf{2.24} & \textbf{2.25} \\

\midrule
w/o decomp & 2.68 & 3.28 & 4.12 & 3.72 \\

simply filter & 3.23 & 3.28 & 3.64 & 3.77 \\

only dc & 5.42 & 4.87 & 6.17 & 5.89 \\
\bottomrule
\end{tabular}
}
\vspace{-0.2cm}
\caption{Reconstruction of high-frequency components on TACO~\cite{liu2024tacobenchmarkinggeneralizablebimanual} dataset. The best in each column is highlighted in bold.}
\vspace{-0.5cm}
\label{tab:recon}
\end{table}
\textbf{Experiment 2:} We test the semantic quality of synthesized motions on the four test splits of TACO. The experimental setting is the same as Table 1 in the main text. Results are shown in Table ~\ref{tab:dcac_semantics}.  We can clearly see that frequency decomposition enhances the semantic quality of the synthesized motion, and this can not be achieved by merely removing too high-frequency noise.
\begin{table}[t]
\centering
\footnotesize
\addtolength{\tabcolsep}{-3pt}
{
\begin{tabular}{c|c c c c|c c c c}
\toprule  
\multirow{2}{*}{Method} &  \multicolumn{4}{c|}{FID ($\downarrow$)} & \multicolumn{4}{c}{RA (\%, $\downarrow$)}\\
\cline{2-9}
 & Test1 & Test2 & Test3 & Test4 & Test1 & Test2 & Test3 & Test4\\
\midrule
MACS~\cite{shimada2023macsmassconditioned3d} & 10.56 & 23.24 & 32.18 & 42.37 & 58.40 & 53.08 & 33.00 & 19.02 \\

DiffH$_{2}$O~\cite{christen2024diffh2odiffusionbasedsynthesishandobject} & 4.34 & 17.04 & 24.92 & 39.20 & 61.40 & 56.70 & 43.67 & 28.15 \\

\textbf{SyncDiff (Ours)} & \textbf{2.70} & \textbf{2.68} & \textbf{22.96} & \textbf{30.23} & \textbf{73.28} & \textbf{85.92} & \textbf{46.90} & \textbf{40.12} \\

\midrule
w/o decomp & 6.44 & 21.21 & 28.67 & 49.58 & 56.60 & 51.85 & 40.02 & 22.18 \\

simply filter & 7.87 & 20.38 & 29.35 & 37.69 & 54.71 & 53.08 & 39.66 & 22.34 \\

only dc & 42.83 & 54.62 & 88.47 & 85.02 & 30.26 & 31.44 & 25.86 & 19.74\\
\bottomrule
\end{tabular}
}
\vspace{-0.2cm}
\caption{Semantic quality of ablation studies on TACO~\cite{liu2024tacobenchmarkinggeneralizablebimanual} dataset. The best in each column is highlighted in bold.}
\vspace{-0.5cm}
\label{tab:dcac_semantics}
\end{table}

\section{Eliminated Details in the Main Text}
\label{sec:C}
\subsection{Formulas for rel and comb}
\label{sec:C.1}
In Sections~\ref{sec:3.5},~\ref{sec:3.6}, for two distinct bodies \textbf{a} and \textbf{b}, we define $\text{rel}(x_{\textbf{a}}, x_{\textbf{b}})$ as $\textbf{b}$'s motion relative to $\textbf{a}$, and let $\text{comb}(x_{\textbf{a}}, x_{\textbf{b}\rightarrow \textbf{a}})$ utilize the individual motion of $\textbf{a}$ and relative motion between $\textbf{b}$ and $\textbf{a}$ to compute $\textbf{b}$'s motion. The detailed expressions are shown below:

If $\textbf{a}=o_{j_1}$ and $\textbf{b}=o_{j_2}$ are two rigid objects, whose motions are $x_{o_{j_1}}=[\textbf{t}_{o_{j_1}}, \textbf{q}_{o_{j_1}}]$ and $x_{o_{j_2}}=[\textbf{t}_{o_{j_2}}, \textbf{q}_{o_{j_2}}]$, denote $x_{o_{j_2}\rightarrow o_{j_1}}=[\textbf{t}_{o_{j_2}\rightarrow o_{j_1}}, \textbf{q}_{o_{j_2}\rightarrow o_{j_1}}]$, then $\text{rel}(x_{o_{j_1}}, x_{o_{j_2}})=\left[\textbf{q}_{o_{j_1}}^{-1}(\textbf{t}_{o_{j_2}}-\textbf{t}_{o_{j_1}}), \textbf{q}_{o_{j_1}}^{-1}\textbf{q}_{o_{j_2}}\right]$, $\text{comb}(x_{o_{j_1}}, x_{o_{j_2}\rightarrow o_{j_1}})=[\textbf{q}_{o_{j_1}}\textbf{t}_{o_{j_2}\rightarrow o_{j_1}}+
\textbf{t}_{o_{j_1}}, \textbf{q}_{o_{j_1}}\textbf{q}_{o_{j_2}\rightarrow o_{j_1}}]$.

If $\textbf{a}=o_j$ is rigid object, and $\textbf{b}=h_i$ is an articulated skeleton, let $x_{o_{j}}=[\textbf{t}_{j}, \textbf{q}_{j}]$, and denote the motion for one of the joints in $x_{h_i}$ as $\textbf{p}_{h_i}$, the motion for one of the joints in $x_{h_i\rightarrow o_j}$ as $\textbf{p}_{h_i\rightarrow o_j}$. We have $\text{rel}(x_{o_{j}}, \textbf{p}_{h_i})=\textbf{q}_{j_1}^{-1}(\textbf{p}_{h_i}-\textbf{t}_{j_1})$, $\text{comb}(x_{o_{j}}, \textbf{p}_{h_i\rightarrow o_{j}})=\textbf{q}_{o_{j}}\textbf{p}_{h_i\rightarrow o_{j}}+\textbf{t}_{o_{j}}$.

Note that $\textbf{a}$ must be a rigid object, in order to define a coordinate system based on its transformation matrix. Also, $\text{rel}\left(\hat{x}_{\textbf{a}}, \hat{x}_{\textbf{b}}\right)$ is different from $\hat{x}_{\textbf{b}\rightarrow \textbf{a}}$. The former is calculated based on the predicted motion of $\textbf{a}$ and $\textbf{b}$, which should approach $\hat{x}_{\textbf{b}\rightarrow \textbf{a}}$ in order to foster synchronization. The latter is directly predicted by the diffusion model, which should fit the distribution of ground-truth $x_{\textbf{b}\rightarrow \textbf{a}}$, in order to increase data fidelity. Similar arguments hold for comb.

\subsection{Post Process: Mesh Reconstruction}
\label{sec:C.2}

For tasks such as robot learning, the joint information for articulated skeletons (hands/humans) is already sufficient. Post processing methods like Reinforcement Learning (RL) or Imitation Learning (IL) can further guide the model to generate physically realistic data with the help of physics-based simulations. Therefore, in such scenarios, we can direct use $\hat{x}_{h_{i\in [1, n]}}$ as predicted results. However, for tasks like animation production or VR/AR, it is necessary to reconstruct the full meshes. In this section, we introduce how to reconstruct MANO hand mesh or SMPL-X human body mesh of natural shape from the predicted joint positions $\mathbb{R}^{N\times D\times 3}$, where $N$ and $D$ are the number of frames and joints.

\textbf{MANO Hand Mesh Reconstruction.} For one single hand, suppose the predicted joint positions are $\hat{x}\in \mathbb{R}^{N\times 21\times 3}$. We start with tunable MANO~\cite{Romero_2017} parameters for joint pose $\theta\in \mathbb{R}^{N\times 45}$, global orientation $R\in \mathbb{R}^{N\times 3}$ and translation $l\in \mathbb{R}^{N\times 3}$, which are all set to zero tensors initially. We freeze hand shape $\beta\in\mathbb{R}^{10}$. During the optimizing process, let the current calculated joint positions be $K\in \mathbb{R}^{N\times 21\times 3}$ based on $\theta$, $R$, and $l$. We want to minimize
$$\mathcal{L}_{\text{MANO}}=\lambda_{\text{pos}}\mathcal{L}_{\text{pos}}+\lambda_{\text{angle}}\mathcal{L}_{\text{angle}}+\lambda_{\text{vel}}\mathcal{L}_{\text{vel}}$$
Here
$$\mathcal{L}_{\text{pos}}=\sum\limits_{t=1}^{N}\sum\limits_{i=1}^{21}\Vert \hat{x}_{t, i}-K_{t, i}\Vert_2^2$$
which minimizes the difference between joint positions $K$ from MANO calculation and our predicted results $\hat{x}$.

$$\mathcal{L}_{\text{angle}}=\sum\limits_{t=1}^{N}\sum\limits_{i=1}^{45} [\max(\theta_{t, i}-u_i, 0)+\max(d_i-\theta_{t, i}, 0)]$$
where $(d_i, u_i)$ is the permitted rotation range for the $i$-th degree of freedom. This loss item ensures that the rotation angle of each joint remains within permittable limits, preventing unreasonable distortions.

$$\mathcal{L}_{\text{vel}}=\sum\limits_{t=1}^{N-1} \Vert l_t-l_{t+1}\Vert_2^2$$
which keeps the positions at adjacent timesteps close to ensure a smooth trajectory without abrupt changes.

\textbf{SMPL-X Human Body Mesh Reconstruction.} Similar to MANO in hand pose representation, for human body pose representation, we also have SMPL-X~\cite{pavlakos2019expressivebodycapture3d} representation. It is composed of a set of parameters $(\theta, R, l, \theta_{\text{left}}, \theta_{\text{right}})$, where the body pose $\theta \in \mathbb{R}^{N \times 21 \times 3}$ represents the $21$ joint orientations(except root), $R, l\in \mathbb{R}^{N\times 3}$ are global orientation and translation, and $\theta_{\text{left}}, \theta_{\text{right}}\in \mathbb{R}^{N\times 12}$ are compressed representations of two hands. Human shape parameters $\beta\in \mathbb{R}^{10}$ are given, while $\hat{x}\in \mathbb{R}^{N\times 22\times 3}$ denotes the joint positions we predict.

Due to the flexibility of human body poses, to reconstruct the human body mesh, we not only need joint positions but also joint orientations. Therefore, we need to introduce additional $N \times 21 \times 3$ degrees of freedom to every $x_{h_i}(i\in [1, n])$, representing the predicted human body pose $\hat{\theta}$ under SMPL-X. These degrees of freedom are also predicted by the diffusion model, but they only participate in the decomposition process and do not get involved in our two synchronization mechanisms.

Similar to hand mesh reconstruction, we start with tunable SMPL-X parameters $R, l, \theta_{\text{left}}, \theta_{\text{right}}$, which are all set to zero tensors. We freeze hand shape $\beta\in\mathbb{R}^{10}$ and $\theta=\hat{\theta}$. During the optimizing process, let the current calculated joint positions be $K\in \mathbb{R}^{T\times 22\times 3}$ based on $R, l, \theta_{\text{left}}, \theta_{\text{right}}$. We want to minimize

$$\mathcal{L}_{\text{SMPL-X}}=\lambda_{\text{pos}}\mathcal{L}_{\text{pos}}+\lambda_{\text{vel}}\mathcal{L}_{\text{vel}}$$

Here
$$\mathcal{L}_{\text{pos}}=\sum\limits_{t=1}^{T}\sum\limits_{i=1}^{22}\Vert \hat{x}_{t, i}-K_{t, i}\Vert_2^2$$
which minimizes the difference between joint positions from SMPL-X calculation and our predicted results.

$$\mathcal{L}_{\text{vel}}=\sum\limits_{t=1}^{T-1} \Vert l_t-l_{t+1}\Vert_2^2$$
which keeps the trajectory smooth. 

The hyperparameters involved in the mesh reconstruction process can be found in Table ~\ref{tab:hyperparameter_hand_mesh} and ~\ref{tab:hyperparameter_human_mesh}.

\begin{table}[h!]
    \centering
    \begin{tabular}{c c c c}
        \toprule
        \textbf{Parameter}       & TACO & OAKINK2 & GRAB \\
        \midrule
        Optimizer      & \multicolumn{3}{c}{AdamW, $\text{lr}=0.01$} \\
        
        Epoch    &    $5$k    &      $8$k   &        $5$k   \\
        
        $\lambda_{\text{pos}}$ & $1$ & $1$ & $1$\\
        
        $\lambda_{\text{angle}}$ & $0.2$ & $0.2$ & $0.05$\\
        
        $\lambda_{\text{vel}}$ & $0.03$ & $0.03$ & $0.02$\\
        \bottomrule
    \end{tabular}
    \caption{Hyperparameters for MANO hand mesh reconstruction.}
    \label{tab:hyperparameter_hand_mesh}
\end{table}

\begin{table}[h!]
    \centering
    \begin{tabular}{c c c}
        \toprule
        \textbf{Parameter}       & CORE4D & BEHAVE\\
        \midrule
        Optimizer      & \multicolumn{2}{c}{AdamW, $\text{lr}=0.001$} \\
        
        Epoch    &    $5$k & $2$k\\
        
        $\lambda_{\text{pos}}$ & $3$ & $1$\\
        
        $\lambda_{\text{vel}}$ & $0.1$ & $0.1$\\
        \bottomrule
    \end{tabular}
    \caption{Hyperparameters for SMPL-X human body mesh reconstruction.}
    \label{tab:hyperparameter_human_mesh}
\end{table}

\subsection{A Brief Introduction to BPS Algorithm}
\label{sec:C.3}
As is discussed in the main text, we use the Basis Point Set (BPS)~\cite{prokudin2019efficientlearningpointclouds} algorithm to encode the geometric features of rigid bodies. Compared to pretrained models like PointNet~\cite{qi2017pointnetdeeplearningpoint} and PointNet++~\cite{qi2017pointnetdeephierarchicalfeature}, BPS is more lightweight and compact. It does not rely on any data-driven methods and places greater emphasis on object surface features, which is crucial in our relative motion synthesis.

Its working principle is as follows: First, a large enough sphere is chosen such that when its center coincides with any object's centroid, it can fully contain the object. In practice, we choose radius $r=1\text{m}$. Then, $1024$ points are randomly sampled from the sphere's volume. BPS representation is computed by calculating the difference from each sampled point to the nearest point on the object's surface. This results in a vector of size $\mathbb{R}^{1024\times 3}$.

\subsection{Details for Ablation Studies}
\label{sec:C.4}
In our experiments in Section~\ref{sec:4}, there are three categories of ablation studies: ``w/o decompose", ``w/o $\mathcal{L}_{\text{align}}$", and ``w/o exp sync".

``w/o decompose" means that we direct concatenate the condition vector with $x_{t}$, which substitute the position of $x_{t, \text{dc}}$ in the pipeline figure, and eliminate the branch of $x_{\text{F}}$ or $x_{\text{ac}}$. Note that $x_t\neq x_{t, \text{dc}}+x_{t, \text{ac}}$, as $x_t$ includes the components with frequencies higher than $\phi_L/2\pi$, where $\phi_L=L/N$, and $L$ is the cutoff boundary.

``w/o $\mathcal{L}_{\text{align}}$" means that we eliminate the term $\lambda_{\text{align}}\mathcal{L}_{\text{align}}$ in total loss.

``w/o exp sync" means that we perform normal denoising steps every time, without explicit synchronization steps. This can be equivalently interpreted as $s=+\infty$.

\subsection{Pseudocode for CSR, CRR, and CSIoU}
\label{sec:C.5}
In this section, we will provide a detailed description for three contact-based metrics: CSR, CSIoU, and CRR.

First, we define two types of contact: surface contact and root contact. Here $o$ is one single object mesh sequence of size $\mathbb{R}^{N \times M \times 3}$, where $M$ is the number of vertices on its mesh. $h$ is hand mesh sequence of size $\mathbb{R}^{N\times 778\times 3}$ in Contact\_Surface, while it denotes trajectories of root joints of two hands of size $R^{N \times 2 \times 3}$ in Contact\_Root, as shown in Algorithm~\ref{alg:contact}.

\begin{algorithm}[h]
\caption{Contact Definitions}
\label{alg:contact}
\resizebox{\linewidth}{!}{
\begin{minipage}{\linewidth}
\begin{algorithmic}[1]
\Procedure{$\text{Contact\_Surface}$}{$o$, $h$}
    \State $\mathbf{c} \gets \mathbf{0}$
    \For{$t \gets 1$ \textbf{to} $T$}
        \State $d \gets \min\limits_{v_1 \in [1, M], v_2 \in [1, 778]} \Vert o_{t, v_1}- h_{t, v_2}\Vert_2$
        \If{$d \leq 5\text{mm}$}
            \State $\mathbf{c}_t \gets 1$
        \EndIf
    \EndFor
    \State \Return $\mathbf{c}$
\EndProcedure

\Procedure{$\text{Contact\_Root}$}{$o$, $h$}
    \State $\mathbf{c} \gets \mathbf{0}$
    \For{$t \gets 1$ \textbf{to} $T$}
        \State $d_1 \gets \min\limits_{v \in [1, M]} \Vert o_{t, v}- h_{t, 1}\Vert_2$
        \State $d_2 \gets \min\limits_{v \in [1, M]} \Vert o_{t, v}- h_{t, 2}\Vert_2$
        \If{$\max(d_1, d_2) \leq 3\text{cm}$}
            \State $\mathbf{c}_t \gets 1$
        \EndIf
    \EndFor
    \State \Return $\mathbf{c}$
\EndProcedure
\end{algorithmic}
\end{minipage}
}
\end{algorithm}

Based on these contact definitions, it comes to the calculation of the three metrics. Here $o$ is the object mesh sequence list of length $m$, and $h$ is the hand mesh sequence or human hand root joint sequence list of length $n$. $o^{'}$ and $h^{'}$ are corresponding ground-truth versions. The calculation is shown in Algorithm~\ref{alg:metrics}.

\begin{algorithm}[h]
\caption{Metric Calculation}
\label{alg:metrics}
\resizebox{\linewidth}{!}{
\begin{minipage}{\linewidth}
\begin{algorithmic}[1]
\Procedure{$\text{CSR}$}{$o$, $h$}
    \State $\text{CSR} \gets 0$
    \For{$i \gets 1$ \textbf{to} $n$}
        \State $\mathbf{c} \gets \mathbf{0}$
        \For{$j \gets 1$ \textbf{to} $m$}
            \State $\mathbf{c} \gets \mathbf{c} \lor \text{Contact\_Surface}(o_j, h_i)$
        \EndFor
        \State $\text{CSR} \gets \text{CSR} + \frac{1}{T} \sum\limits_{t=1}^T \mathbf{c}_t$
    \EndFor
    \State \Return $\text{CSR}/n$
\EndProcedure

\Procedure{$\text{CSIoU}$}{$o$, $h$, $o^{'}$, $h^{'}$}
    \State $\text{CSIoU} \gets 0$
    \For{$i \gets 1$ \textbf{to} $n$}
        \State $\mathbf{c1} \gets \mathbf{0}, \mathbf{c2} \gets \mathbf{0}$
        \For{$j \gets 1$ \textbf{to} $m$}
            \State $\mathbf{c}_1 \gets \mathbf{c}_1 \lor \text{Contact\_Surface}(o_j, h_i)$
            \State $\mathbf{c}_2 \gets \mathbf{c}_2 \lor \text{Contact\_Surface}(o^{'}_j, h^{'}_i)$
        \EndFor
        \State $\text{CSIoU} \gets \text{CSIoU} + \text{IoU}(\mathbf{c}_1, \mathbf{c}_2)$
    \EndFor
    \State \Return $\text{CSIoU}/n$
\EndProcedure

\Procedure{$\text{CRR}$}{$o$, $h$}
    \State $\text{CRR} \gets 0$
    \For{$i \gets 1$ \textbf{to} $n$}
        \State $\mathbf{c} \gets \mathbf{0}$
        \For{$j \gets 1$ \textbf{to} $m$}
            \State $\mathbf{c} \gets \mathbf{c} \lor \text{Contact\_Root}(o_j, h_i)$
        \EndFor
        \State $\text{CRR} \gets \text{CRR} + \frac{1}{T} \sum\limits_{t=1}^T \mathbf{c}_t$
    \EndFor
    \State \Return $\text{CRR}/n$
\EndProcedure
\end{algorithmic}
\end{minipage}
}
\end{algorithm}

\section{Important Statistics}
\label{sec:D}
\subsection{Dataset Statistics}
\label{sec:D.1}
The sizes of each dataset split are shown in Table ~\ref{tab:dataset_statistics}.

\begin{table}[ht]
    \centering
    \begin{tabular}{l|l}
        \toprule
        \textbf{Dataset}       & \textbf{Statistics}                     \\ \midrule
        TACO               & train:test1:test2:test3:test4\\ & =1035:238:260:403:610                                \\ 
        \hline
        CORE4D & train:test1:test2=483:197:195                                     \\ 
        \hline
        OAKINK2      & train:val:test=1884:167:723\\
        \hline
       GRAB & train:val:test=992:198:144 (Unseen Subject) \\
        & train:test=1126:208 (Unseen Object) \\
        \hline
        BEHAVE & train:test=231:90 \\ \bottomrule
    \end{tabular}
    \vspace{-0.2cm}
    \caption{Dataset statistics.}
    \vspace{-0.2cm}
    \label{tab:dataset_statistics}
\end{table}

\subsection{Hyperparameters in Model Architecture}
\label{sec:D.2}

\textbf{Action/object Label Feature Extraction Branch.} Action/object label features are first encoded by pretrained CLIP~\cite{radford2021learningtransferablevisualmodels}, and then pass through a 2-layer MLP. Specific parameters are shown in Table ~\ref{tab:clip_config}.

\begin{table}[h!]
    \centering
    \begin{tabular}{l l}
        \toprule
        \textbf{Component}       & \textbf{Description}                     \\ \midrule
        CLIP Type               & ViT-B/32                                \\ 
        \midrule
        Raw Feature Space Dimension & $512$                                     \\ 
        \midrule
        MLP Architecture      & Linear($512$, $512$) \\
        & ReLU() \\
        & Linear($512$, $128$) \\ \bottomrule
    \end{tabular}
    \caption{Hyperparameters in label feature extraction branch.}
    \label{tab:clip_config}
\end{table}

\textbf{Object Geometry Feature Extraction Branch.} Object geometry features are first encoded by pretrained BPS~\cite{prokudin2019efficientlearningpointclouds}, and then pass through a 2-layer MLP. Specific parameters are shown in Table ~\ref{tab:bps_config}.
\begin{table}[ht!]
    \centering
    \begin{tabular}{l l}
        \toprule
        \textbf{Component}       & \textbf{Description} \\            
        \midrule
        Raw Feature Space Dimension & $1024\times 3 $                                   \\ 
        \midrule
        MLP Architecture      & Linear($1024\times 3$, $512$) \\
        & ReLU() \\
        & Linear($512$, $128$) \\ \bottomrule
    \end{tabular}
    \caption{Hyperparameters in rigid body geometry feature extraction branch.}
    \label{tab:bps_config}
\end{table}

\textbf{Noise Timestep Embedding Module.} Detailed architecture is shown in Table ~\ref{tab:time_embed}.
\begin{table}[ht!]
    \centering
    \begin{tabular}{l l}
        \toprule
        \textbf{Component}       & \textbf{Description} \\            
        \midrule
        Timestep Embedder &
        SinusoidalPosEmbedding($64$)   \\
        & Linear($64$, $256$) \\
        & GeLU() \\
        & Linear($256$, $1024$) \\ \bottomrule
    \end{tabular}
    \caption{Hyperparameters in noise timestep embedding.}
    \label{tab:time_embed}
\end{table}

\textbf{Transformer Encoder-Decoder.} Concatenate action label features, object label features (for $m$ rigid objects), object geometry features (for $m$ rigid objects), and the shape parameters $\beta$ of $n$ articulated skeletons to form a condition vector, whose shape is $\mathbb{R}^{128\times (2m+1)+10n}$, as is mentioned in ~\ref{sec:3.4}. After replicating and concatenating with padding mask, $x_{\text{dc}}$ and $x_{\text{F}}$, the dimension becomes $\mathbb{R}^{N\times \left(128×(2m+1)+10n+7m+3Dn+7m(m-1)+3Dmn+1\right)}=\mathbb{R}^{N\times \left(128×(2m+1)+10n+D_{\text{sum}}+1\right)}=\mathbb{R}^{N\times \mathcal{C}}$. The architecture of the transformer encoder-decoder is shown in Table ~\ref{tab:transformer_config}.

\begin{table}[h!]
    \centering
    \begin{tabular}{l l}
        \toprule
        \textbf{Component}       & \textbf{Description} \\            
        \midrule
        Encoder & Conv1D($\mathcal{C}$, $512$)                                   \\ 
        
        Latent Transformer      & $4$-layer, $8$-head, $1024$-dim \\
        
        Decoder & Linear($512$, $D_{\text{sum}}$) \\
        \bottomrule
    \end{tabular}
    \caption{Hyperparameters in the transformer encoder-decoder.}
    \label{tab:transformer_config}
\end{table}

Note that after $x_{\text{dc}}$ and $x_{\text{F}}$ (concatenating with the condition vector) passes through the encoder, their shapes becomes $N\times 512$. Concatenating them together results in a latent vector of shape $N\times 1024$, which can be affiliated by the noise embedding, whose dimension is $1024$.

\subsection{Training and Inference Hyperparameters}
\label{sec:D.3}
Other important hyperparameters for training and inference process are shown in Table ~\ref{tab:hyperparameter}.

\begin{table}[ht!]
    \centering
    \footnotesize
    \begin{tabular}{c c c c c c }
        \toprule
        \textbf{Parameter}       & TACO & CORE4D & OAKINK2 & GRAB & BEHAVE\\            
        \midrule
        Batch Size & \multicolumn{5}{c}{$32$}                                   \\ 
        
        Optimizer      & \multicolumn{5}{c}{Adam, $\text{lr}=0.0001$, $\text{ema\_decay}=0.995$} \\
        
        Epoch    &    $250$k    &      $140$k   &        $280$k    &         $100$k & $100$k\\
        
        $\lambda_{\text{DC}}$ & $1$ & $1$ & $1$ & $1$ & $1$\\
        
        $\lambda_{\text{AC}}$ & $2.5$ & $0.8$ & $0.8$ & $0.3$ & $0.8$\\
        
        $\lambda_{\text{norm}}$ & $0.1$ & $0.1$ & $0.1$ & $0.1$ & $0.1$\\
        
        $\lambda_{\text{align}}$ & $0.3$ & $0.3$ & $0.3$ & $0.15$ & $0.15$\\
        
        $\lambda_{\text{exp}}$ & $0.3$ & $0.3$ & $0.3$ & $0.3$ & $0.3$\\
        
        Diffusion & \multicolumn{5}{c}{$\alpha\in [0.0001, 0.01]$, Uniform}\\
        \bottomrule
    \end{tabular}
    \vspace{-0.2cm}
    \caption{Hyperparameters for training and inference process in SyncDiff.}
    \vspace{-0.2cm}
    \label{tab:hyperparameter}
\end{table}

\subsection{Time and Space Cost, Hardware Configurations}
We conduct experiments on NVIDIA A40. All operations can be performed on a single GPU.

\textbf{Time and Space Cost for Training and Inference.} The training time, average inference time per sample, and GPU memory usage during training are detailed in Table ~\ref{tab:timecost}.

\begin{table}[h!]
    \centering
    \footnotesize
    \begin{tabular}{c c c c c c}
        \toprule
        \textbf{Cost}       & TACO & CORE4D & OAKINK2 & GRAB & BEHAVE\\            
        \midrule
        Training & $20.7$h & $9.5$h & $40.1$h & $7.9$h & $3.7$h                                   \\ 
        
        Inference      & $7.7$s & $6.5$s & $7.3$s & $6.6$s & $4.2$s\\
        
        Memory    &    $6.11$G & $5.07$G & $8.59$G & $4.93$G & $3.67$G\\
        \bottomrule
    \end{tabular}
    \vspace{-0.2cm}
    \caption{Time and space costs for training and inference process in SyncDiff on different datasets.}
    \vspace{-0.2cm}
    \label{tab:timecost}
\end{table}

\textbf{Time and Space Cost for Mesh Reconstruction.} Time and space cost of performing mesh reconstruction for a motion sequence of $N=200$ frames are shown in Table ~\ref{tab:timecost_mesh}. 
\begin{table}[h]
    \centering
    \footnotesize
    \begin{tabular}{ c c c c c c}
        \toprule
        \textbf{Cost}       & TACO & CORE4D & OAKINK2 & GRAB & BEHAVE \\            
        \midrule
        Time & $130$s & $144$s & $206$s & $130$s & $74$s                                   \\ 
        
        Memory   &    $492$M & $826$M & $492$M & $492$M & $430$M\\
        \bottomrule
    \end{tabular}
    \vspace{-0.2cm}
    \caption{Time and space costs for mesh reconstruction in SyncDiff on different datasets.}
    \vspace{-0.2cm}
    \label{tab:timecost_mesh}
\end{table}

Although the mesh reconstruction operation seems to take much more time than the inference process, due to parallelized calculation, the amortized time complexity is relatively low. In addition, the mesh reconstruction process is also optional, since in tasks like robot planning, only joint positions are enough.

\section{Limitations and Discussions}

\subsection{Current Limitations and Potential Solutions}
Some current limitations of SyncDiff and their potential solutions are as follows:

\textbf{1. Lack of Articulation-Aware Modeling.} Our method models articulated objects (such as those in OAKINK2~\cite{zhan2024oakink2datasetbimanualhandsobject}) as part-wise rigid body individuals directly and coordinates their motions without leveraging their intrinsic articulations. Integrating these articulations into multi-body likelihood modeling could be an interesting future direction. The potential solution may be unify them with the articulated skeletons (hands/humans), and treat their intrinsic articulations as conditions.

\textbf{2. High-cost of Explicit Synchronization Step.} As body number increases, the time consumption for the calculation of alignment loss and explicit synchronization step grows quadratically. Note that for multi-body HOI synthesis, not all pairwise relationships are necessary. A possible solution is to use human priors or another algorithm to filter out the relationships that truly require synchronization, and perform synchronization only across them.

\textbf{3. Lack of Physically Accurate Guarantees.} Unlike methods that utilize true physical simulations, our approach cannot guarantee physical truthfulness. In many cases, minor errors can be observed in the supplementary videos, but these small discrepancies may be sufficient to cause visible failures in real tasks. In robot manipulation tasks, we prefer to treat SyncDiff as a robot planning method that requires downstream integration with physically accurate optimization (RL/IL) to ensure practical usability in real-world applications.



\textbf{4. Limitations in Pure Multi-human Interaction Synthesis.} Since the relative representations need to be generated in the coordinate systems of rigid bodies, whose motions can be represented by translations and rotations, our method may not be directly adapted for pure multi-human interaction synthesis. To address this limitation, more complex relative representations is required. A potential solution is use representations similar to MANO~\cite{Romero_2017}, representing the motions of an articulated skeleton as the transformation matrix and local relative positions of its joints. Although this approach sacrifices the homogeneity of the articulated skeleton motion representation, it allows for a more convenient computation of the relative representation between two skeletons. This idea is manifested in a bunch of multi-human interaction synthesis works like ComMDM~\cite{shafir2023humanmotiondiffusiongenerative} and InterGen~\cite{Liang_2024}.

\subsection{Discussions and Some Extensions}
\label{sec:E}
This section aims to enumerate some points where readers may have questions or misconceptions, providing detailed explanations. For scenarios where SyncDiff can be extended, we will also offer intuitive methods for expansion.

\textbf{1. Does SyncDiff lose the flexibility of ``one model for all datasets"?}

Due to the scarcity of mocap datasets, as well as the need to adopt models into multiple scenarios, sometimes it is necessary to merge data from multiple datasets for training. However, due to the fixed number of parameters, a trained model of a specific size can only handle a pair of (number of rigid objects, number of articulated skeletons). For instance, the data from TACO can only be merged with the 2-hand 2-object samples in OAKINK2 for training, but not directly with GRAB.

\textbf{Our response is divided into three aspects.}

First, the extension from a unified framework to ``one model for all datasets" is basically trivial, only increasing some computational load. It is noted that any graphical model is a subgraph of a sufficiently large complete graph. By using masks to cover the individual bodies that do not need to be synthesized and the relative relations that do not need to be adopted, this can be achieved. Such an idea is also reflected in ~\cite{shan2025towards}.

Second, existing sota methods can only train for specific (number of rigid objects, number of articulated skeletons), and even require the type of articulated skeleton to be given (for example, the carefully designed grasp guidance in DiffH2O is almost ineffective when applied to the human-object interaction dataset CORE4D; the cross-attention between bodies and contact maps in CG-HOI is too coarse-grained for dexterous hand-object interaction). Our method, as a unified framework, significantly reduces the cost of manually trying different pipelines and designing different guidance/representations for different settings, which is a giant leap forward.

Third, even if our method is treated as a single-track method for every specific configuration, it surpasses many sota methods in the corresponding tracks. As body number increases, motion semantics become more complex, especially between multiple rigid objects. For example, the object trajectories in the GRAB/BEHAVE dataset mainly consist of basic units like picking up, putting down, and lateral movements. However, in TACO, there may be actions such as rubbing back and forth, tapping, and pouring between two items. If OAKINK2 involves modeling the rigid parts of articulated objects, it also needs to ensure coordination between these different parts, such as a test tube passing precisely between the two halves of a test tube holder. As the first work to adopt synchronization and frequency decomposition to model complex motions, we have also significantly outperformed existing state-of-the-art methods in settings with a larger number of bodies.

\textbf{2. Why doesn't SyncDiff use the 6 DoF rotation representation that is widely employed in the current motion synthesis methods?}

To represent the rotation $R\in SO(3)$ of an object, the 6 DoF representation proposed in ~\cite{zhou2019on} concatenates the first two columns of the $3\times 3$ rotation matrix into a 6-dimensional vector $r$ as the representation. After the model generates the predicted result $\hat{r}$, it is split into two column vectors, normalized, and subjected to Gram-Schmidt orthogonalization, and then completed into the rotation matrix $\hat{R}$ in $SO(3)$.

In our method, since the relative representations need to be solved frequently, if we use such 6 DoF representation, it is necessary to complete them into matrices for operations such as inversion and composition, which significantly increases the computational cost. Therefore, we adopt quaternions, which are more convenient for normalization and calculation.

\textbf{3. Is the explicit synchronization during inference in SyncDiff inspired by the guidance strategy in Guided Motion Diffusion (GMD)~\cite{karunratanakul2023guided}?}

As is stated in Section~\ref{sec:2.2}, several prominent works inject external priors or constraints into synthesized results by performing linear fusions, imputations or inpaintings during the inference process of diffusion models. The term $-\nabla _{\textbf{x}_0}G_z(P_x^z\textbf{x})$ in GMD and the gradient of negative log-likelihood (actually equivalent to the score functions) in SyncDiff (Refer to Section~\ref{sec:A.2}) both provide the direction of denoising (for both), adhering to constraints (for GMD), or synchronization (for SyncDiff).

The distinction lies in that different application scenarios have led to different concrete operations for the same idea. GMD aims to constrain the generation of high-dimensional human pose $x$ controlled by the goal function $G_z$, where $z$ is low-dimensional pelvis trajectory. Hence, it employs an inpainting/imputation strategy to project (by the projector $P_x^z$) and complete the sparse trajectory during inference. In contrast, in SyncDiff, ``synchronization" is a concept that is difficult to quantify. Therefore, we introduce the probabilistic modeling on graphical models, and ensure that synchronization and data denoising are simultaneously established in the inference process.

\textbf{4. Can the articulated skeletons in SyncDiff be other body parts besides hands and the full body?}

In Section~\ref{sec:3.1}, our definition of articulated skeletons is ``a set of joints whose motion can be reconstructed based on joint information and shape parameter $\beta$". To extend SyncDiff to HOI motion synthesis of other body parts, \textbf{two issues need to be addressed:}

1) Collect relevant high-precision mocap data. For example, if we want to perform HOI motion synthesis for feet, we need to collect high-quality data such as kicking a ball, putting on shoes, ice skating, etc.

2) Design a data representation for motion reconstruction based on joint information and shape parameter $\beta$, similar to MANO for hands or SMPL-X for human bodies.

\textbf{5. Does SyncDiff's ignorance of affordance result in the item being manipulated incorrectly? For example, the cup was not picked up from the handle position?}

SyncDiff does not consider affordance because the five datasets we use do not provide affordance for rigid objects at all. If the affordance needs to be considered, the relevant data with affordance map needs to be collected first, and the next potential solution may be to encode it in some way as part of the object's geometric features.

In addition, the data-driven generative model should fit the data distribution. As long as the dataset used for training ensures high-quality and correct interactions in a wide range, the model should learn to generate similar samples with correct manipulation by itself.

\textbf{6. Is the metric RA (Recognition Accuracy) fair? Since the discriminator is trained on the combination of train, val, and test folders, is there any risk of overfitting?}

Training the classifier solely on the training split would introduce a more serious issue: motion synthesis models overfitting to training distribution would achieve inflated RA scores. In fact, experiments reveal that this approach causes ground-truth test splits to underperform some baselines. This is because the distributional difference between the ground-truth test splits and train splits is greater than that between the distribution fitted by the generative model on train split and the train split itself.

The fundamental challenge is that any classifier—regardless of its training data—will inherently favor in-domain motions. While a huge and more diverse dataset could mitigate this bias, it is currently impractical due to the acquisition cost of mocap data. A motion recognition foundation model might help mitigate this issue. However, currently there are no such foundation models capable of performing high-quality action recognition based solely on trajectories rather than RGB video inputs.

Besides, the relative rankings produced by our RA metric (Tables 1-4 in main text, Table ~\ref{tab:table_behave}) align with perceptual judgments in user studies (Figure ~\ref{fig:userstudy}), supporting its validity. 

\end{document}